\documentclass{article} 
\usepackage{iclr2025_conference,times}

\usepackage{amsmath,amsfonts,bm}

\def\eqref#1{equation~\ref{#1}}

\def\1{\bm{1}}

\DeclareMathAlphabet{\mathsfit}{\encodingdefault}{\sfdefault}{m}{sl}
\SetMathAlphabet{\mathsfit}{bold}{\encodingdefault}{\sfdefault}{bx}{n}

\usepackage{url}
\usepackage[utf8]{inputenc} 
\usepackage{xcolor}

\usepackage[utf8]{inputenc} 
\usepackage[T1]{fontenc}    
\usepackage{url}            
\usepackage{booktabs}       
\usepackage{amsfonts}       
\usepackage{nicefrac}       
\usepackage{microtype}      
\usepackage{xcolor}         

\usepackage{amsmath}
\usepackage{graphicx}

\usepackage{booktabs}
\usepackage{multirow}
\usepackage{wrapfig}
\usepackage[hidelinks]{hyperref}

\usepackage{makecell} 

\title{Seeing Through the PRISM: Compound \& \\Controllable Restoration of Scientific Images}

\author{Rupa Kurinchi-Vendhan\thanks{Corresponding author: rupak272@mit.edu}, Pratyusha Sharma, Antonio Torralba, Sara Beery\\
Massachusetts Institute of Technology\\
}

\usepackage{xcolor}

\iclrfinalcopy 
\begin{document}

\maketitle

\begin{abstract}
Scientific and environmental imagery often suffer from complex mixtures of noise related to the sensor and the environment. Existing restoration methods typically remove one degradation at a time, leading to cascading artifacts, overcorrection, or loss of meaningful signal. In scientific applications, restoration must be able to simultaneously handle compound degradations while allowing experts to selectively remove subsets of distortions without erasing important features. To address these challenges, we present \textbf{PRISM} (\textbf{P}recision \textbf{R}estoration with \textbf{I}nterpretable \textbf{S}eparation of \textbf{M}ixtures). PRISM is a prompted conditional diffusion framework which combines compound-aware supervision over mixed degradations with a weighted contrastive disentanglement objective that aligns primitives and their mixtures in the latent space. This compositional geometry enables high-fidelity joint removal of overlapping distortions while also allowing flexible, targeted fixes through natural language prompts. Across microscopy, wildlife monitoring, remote sensing, and urban weather datasets, PRISM outperforms state-of-the-art baselines on complex compound degradations, including zero-shot mixtures not seen during training. Importantly, we show that selective restoration significantly improves downstream scientific accuracy in several domains over standard ``black-box'' restoration. These results establish PRISM as a generalizable and controllable framework for high-fidelity restoration in domains where scientific utility is a priority. Our code and dataset are available at this \href{https://github.com/RupaKurinchiVendhan/PRISM}{\textit{link}}.
\end{abstract}

\vspace{-3mm}
\begin{figure}[h]
    \centering    \includegraphics[width=0.90\textwidth]{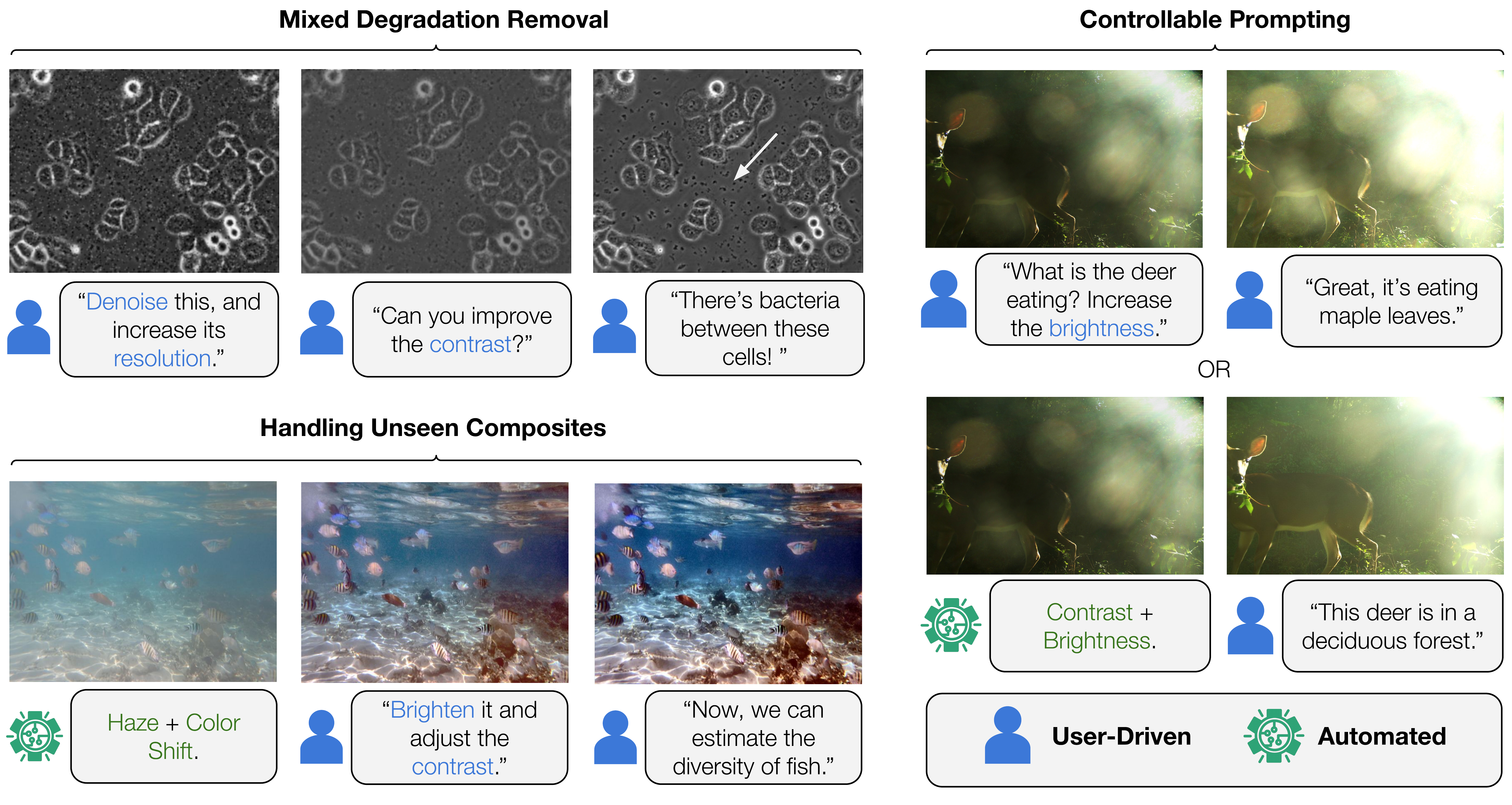}
    \caption{\textit{Expert-in-the-Loop Restoration with PRISM.} PRISM enables robust compound restoration and zero-shot handling of unseen mixtures. It supports both automatic restoration and prompt-driven, selective correction for scientific analysis.}
    \label{fig:motivation}
\end{figure}

\section{Introduction}
\label{sec:intro}

Scientific and environmental imagery is rarely affected by a single degradation. Instead, images typically suffer from \textit{compounding effects} that vary across datasets and collection settings: underwater images combine low light, scattering, and wavelength-dependent absorption effects \citep{akkaynak2018revised, chiang2011underwater}, while satellite imagery suffers from overlapping sensor noise, haze, and cloud occlusions \citep{king2013spatial, ahmad2019haze}.  

Specialized single-distortion models (e.g., for dehazing or removing clouds) enable domain experts to preprocess noisy data before conducting analysis. These approaches are often carefully hand-crafted for specific datasets and distortion types, making them brittle when degradations occur unpredictably, especially in scientific settings where ground truth is unavailable and the underlying distortions are not known a priori. For example, camera trap data may be affected by combinations of motion blur, weather, and lighting that can vary across images from the same deployment, making single-distortion pipelines ineffective. While this has motivated generalist ``all-in-one'' models \citep{li2020all, potlapalli2023promptir}, current frameworks perform sequential/iterative removal of single distortions, which may lead to cascading artifacts and indiscriminately remove errors, erasing signals that should be preserved.

In scientific applications, restoration must preserve signals critical for precision and analysis, not just aesthetics. Aggressive restoration can inadvertently introduce more error: denoising may erase faint galaxies in astronomy data \citep{starck2002deconvolution}, and super-resolution can hallucinate subcellular structures in microscopy images \citep{christensen2022deep}. Few models let experts control these tradeoffs. 

We argue that restoration in scientific contexts requires three principles: \textit{simultaneous over sequential correction}, \textit{precision over aesthetics}, and in some cases, \textit{control over automation}. We introduce PRISM, a conditional diffusion framework that simultaneously disentangles compound degradations and enables expert-guided, precision restoration. Figure \ref{fig:motivation} shows example use-cases of PRISM across scientific domains, demonstrating how our approach enables robust and interactive mixed degradation removal, how automated restoration can infer distortion types from an input image, and how expert prompts can selectively target distortions while preserving critical features. Our contributions are:
\begin{enumerate}
    \item A \textbf{principled embedding design for compound degradations,} combining weighted contrastive learning with compound-aware supervision. The resulting \textbf{structured, compositional latent space} yields separable embeddings for primitives and their mixtures, enabling both automated compound restoration and selective distortion-specific control;
    \item A \textbf{novel benchmark for scientific utility} spanning remote sensing, ecology, biomedical, and urban domains---including our newly-introduced Rooftop Cityscapes dataset---that reflects the needs of scientific workflows by evaluating task fidelity rather than perceptual quality;
    \item A systematic study showing that \textbf{controllability is not a convenience but a necessity}. Selective, distortion-specific guidance significantly improves downstream scientific accuracy under compound degradations.
\end{enumerate}

\section{Related Works}
\label{sec:related_works}

\subsection{Restoration in Scientific Domains}
Restoration has long been important for scientific imaging: early astronomy corrected photographic plates \citep{gull1978image}, while biomedical imaging relied on denoising and deblurring for diagnostics \citep{buades2005image, dabov2007image}. Modern deep learning pipelines destripe astronomical surveys \citep{liu2025astronomical, vojtekova2021learning} and denoise MRI images \citep{yan2024self, manjon2018mri, kidoh2020deep}. While effective within their target domains, these methods typically assume that degradations are fixed, known, and well represented at training time.

In practice, scientific images often exhibit \emph{compound} degradations. Domain-specific models such as Sea-Thru \citep{akkaynak2019sea} for underwater image correction explicitly model these effects, but rely on tailored, paired datasets which can limit their generalization when conditions deviate beyond their assumptions.

Additionally, cleaner data is not always better: scientific restoration often requires targeted improvements rather than blanket restoration. In microscopy, \citet{lu2025unet} show that over-denoising can erase weak but biologically meaningful structures; in underwater monitoring, \citet{cecilia2022denoising} find that generic denoisers oversmooth fine-scale marine features critical for ecological interpretation. These findings motivate restoration frameworks that can handle compound effects while allowing experts to control \emph{which} degradations are addressed.

\subsection{Handling Compound Degradations}
To address compound corruptions, several works propose shared-backbone or ``all-in-one'' architectures that jointly train on multiple degradation types, including All-WeatherNet \citep{li2020all}, TransRestorer \citep{chen2021pre}, and SmartAssign \citep{wang2023smartassign}. More recent universal models such as MT-Restore \citep{chen2022learning}, All-in-OneNet \citep{li2022all}, and PatchDiffuser \citep{ozdenizci2023restoring} improve flexibility by expanding training coverage across degradation categories. These approaches use single-distortion removal for compositional restoration which may introduce unwanted artifacts or propagating errors from one stage to the next. Composite approaches, including OneRestore \citep{guo2024onerestore} and AllRestorer \citep{mao2024allrestorer}, explicitly model interactions between degradations rather than treating them independently, leading to improved performance on benchmark mixtures seen during training. Although these methods improve robustness to mixtures observed during training, they do not explicitly enforce a compositional latent structure that guarantees predictable behavior when selectively conditioning on subsets of degradations.

Research in disentangled representation learning suggests that factorized latent structures significantly improve generalization to novel combinations of known components \citep{burgess2018understanding, lake2017building}, particularly when architectures are designed to maintain the independence of these latent features as they transition through successive layers \citep{liang2025compositional, trauble2021disentangled}. Our approach builds on this theory by aligning degradation mixtures with their primitives in the latent space to enable generalization to unseen data.

\subsection{Conditional Diffusion and Prompt-Guided Restoration}
Closely related methods on blind image restoration (BIR) handle cases where degradation types and severities are unknown at test time. Early GAN-based approaches such as BSRGAN \citep{zhang2021designing} and Real-ESRGAN \citep{wang2021real} simulate diverse degradation pipelines to improve robustness under real-world conditions. Diffusion-based restoration more broadly extends conditional modeling with higher fidelity and stochastic control \citep{ho2020denoising, dhariwal2021diffusion, rombach2022high}. More recent diffusion-based BIR methods, including StableSR \citep{nagar2023adaptation} and DiffBIR \citep{lin2024diffbir}, leverage pretrained diffusion priors to generate realistic and high-fidelity restorations without requiring explicit degradation labels. These models demonstrate strong perceptual quality and robustness to complex degradations. However, they rely on learned image priors and do not explicitly use semantic information from vision-language representations.

Prompt-guided approaches offer a more flexible interface for controlling restoration behavior. Methods such as PromptIR \citep{potlapalli2023promptir}, demonstrate that conditioning on auxiliary text improves all-in-one restoration across a fixed vocabulary of degradation categories. DiffPlugin \citep{liu2024diff}, MPerceiver \citep{ai2024multimodal}, and AutoDIR \citep{jiang2024autodir} introduce text- and image-conditioned diffusion models for universal image restoration. However, in these approaches, degradations are not explicitly disentangled in the model's latent space. As a result, removing one distortion may modify the image in unexpected ways, limiting reliable partial restoration. We distinguish between prompt-conditioned restoration and structurally controllable restoration: the former allows specifying tasks, whereas the latter requires a representation in which degradation factors are disentangled enough to support predictable, selective intervention.

In parallel, several works adapted vision–language models for restoration by fine-tuning CLIP representations to be degradation-aware. DA-CLIP aligns image embeddings with textual degradation descriptions, improving robustness and downstream restoration performance under domain shift \citep{luo2023controlling, luo2024photo}. AutoDIR further introduces a semantic-agnostic loss that encourages CLIP to distinguish clean from degraded images based on quality-related cues rather than semantic content, biasing the encoder toward degradation-sensitive features \citep{jiang2024autodir}. These works primarily focus on aligning representations to individual distortion types rather than enforcing compositional structure that supports systematic generalization to unseen mixtures.

A recent survey by \citet{jiang2025survey} highlights limited support for real-world compound degradations and appropriate evaluation protocols. Existing methods also lack compositional structure for stable joint correction and distortion-specific control. We address these gaps by modeling degradations compositionally and introducing a mixed degradations benchmark and a downstream scientific task evaluation, which together assess structural generalization and practical utility under realistic compound settings.
\section{Methods}
\label{sec:methods}

PRISM combines compound-aware supervision with contrastive latent disentanglement to enable robust and controllable restoration under mixed degradations.

\subsection{Building a Dataset of Compound Degradations}

We construct a synthetic dataset from diverse scientific domains: ImageNet \citep{deng2009imagenet}, Sentinel-2 patches from Sen12MS \citep{schmitt2019sen12ms}, iWildCam 2022 \citep{beery2022iwildcam}, EUVP underwater imagery \citep{islam2020fast}, CityScapes \citep{cordts2016cityscapes}, BioSR microscopy slides \citep{gong2021deep}, Brain Tumor MRI \citep{msoud_nickparvar_2021}, and high-resolution Subaru/HSC sky surveys \citep{miao2024astrosr}. Across these datasets, we sample 2M clean images that serve as the ground truth targets during training. See Appendix \ref{sec:dataset} for more details.

\noindent
\textbf{Compound-Aware Supervision.}
Each image is degraded by up to three distortions sampled from a library including geometric warping, blur, photometric shifts and weather effects, etc. We apply a maximum of three distortions per image to capture challenging compound cases while maintaining enough of the original semantic content (see Table \ref{tab:n_sweep_full} in Appendix \ref{sec:ablations_supp}). Each image is distorted by a composition of multiple augmentations.

Following prior work such as Real-ESRGAN \citep{wang2021real}, which demonstrate that increased variability in training degradations can improve robustness and generalization, we construct a diverse spectrum of degradations to better approximate real-world conditions \citep{luo2024photo, luo2022deep, zhang2021designing}. The distortions are applied in random order with varied parameters, such as kernel size for blurring or angle of snowfall, that determine degradation intensity. We use GPT-4 \citep{hurst2024gpt} to generate variable natural language prompts describing image degradations to simulate realistic user instructions (e.g., ``remove haze,'' ``dehaze this image,'' ``dehaze and super-resolve this''). We also include \emph{partial} prompts (remove a subset of distortions) and \emph{negative} prompts (remove a non-present distortion). This design is critical for selective restoration: by exposing the model to submixtures and negative conditions, we encourage it to associate each degradation primitive with a distinct latent direction and to avoid unintended corrections when a distortion is not specified. This supervision explicitly supports predictable, distortion-specific intervention at inference time. Overall, the dataset consists of triplets $(I_{\text{clean}}, I_{\text{dist}}, p)$. Further details on dataset sampling and the distortion pipeline are provided in Appendix \ref{sec:dataset}.

\subsection{{The PRISM Model}}

\begin{figure}
    \centering
    \includegraphics[width=\textwidth]{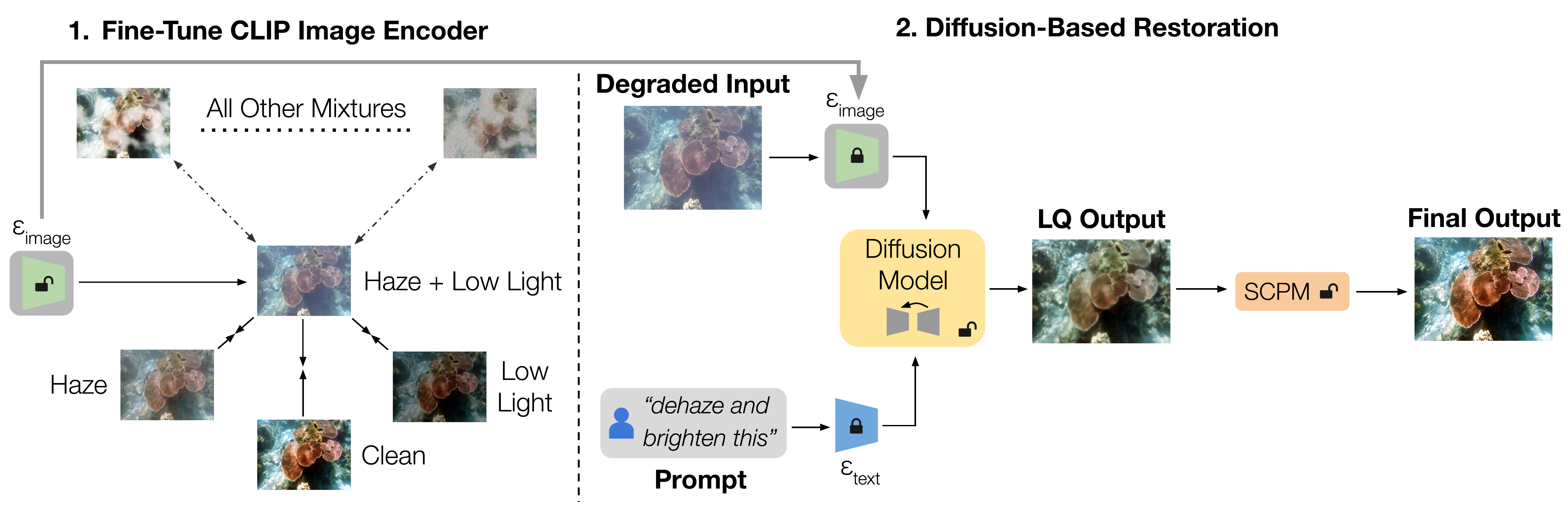}
    \caption{\textit{Overview of PRISM.} We first fine-tune CLIP's image encoder to disentangle image embeddings by distortion. The degraded input and user prompt are then used to condition the latent diffusion backbone, the coarse outputs of which are refined with a Semantic Content Preservation Module (SCPM) to yield the final restored output. Appendix Figure 12 shows the SCPM's architecture.}
    \label{fig:methods}
    \vspace{-4mm}
\end{figure}

Our framework builds on composite/all-in-one restoration \citep{guo2024onerestore, jiang2024autodir, ai2024multimodal} but emphasizes \textit{controllability} and \textit{precision} under compound degradations. Figure \ref{fig:methods} outlines our process: we first fine-tune the CLIP image encoder on our mixed degradation dataset to ensure that embeddings preserve semantic content while becoming distortion-invariant, keeping the text encoder frozen. Once adapted, we freeze both CLIP encoders to provide a stable conditioning space for training the latent diffusion backbone.

\textbf{Stage 1: Disentangling Distortions.}
Naive CLIP embeddings are poorly suited for restoration, as they primarily cluster images by semantic content rather than degradation type or quality \citep{radford2021learning}. Prior work~\citep{jiang2024autodir} showed that quality-aware embeddings improve performance by shifting focus from semantics to distortions. We extend this to compound degradations by explicitly modeling \textit{compositionality}, ensuring embeddings reflect the presence and overlap of multiple distortions (e.g., an image degraded by haze+rain should be more similar to haze-only and rain-only images than to images degraded by unrelated distortions such as noise).

Let $I_{\text{clean}}$ be a clean image and {$\{I_{\text{dist}}^{(j)}\}_{j=1}^{m}$ its $m$ degraded variants}. For cosine similarity $\mathrm{sim}(\cdot, \cdot)$, we use a contrastive loss aligning the corresponding embeddings $e_{\text{dist}}^{(j)}$ with $e_{\text{clean}}$ and repelling it from its $m-1$ sibling variants and all variants from other images in the batch. To encode the similarity structure of different combinations of distortions, we use the Jaccard distance \citep{jaccard1901distribution} between their distortion sets $d^{(j)}$ and $d^{(k)}$
{\[
w_{jk} = \exp\left(1 - \frac{|d^{(j)} \cap d^{(k)}|}{|d^{(j)} \cup d^{(k)}|}\right).
\]}

We analyze this weighting mechanism in Appendix \ref{sec:ablations_supp}. The resulting per-variant loss is
\[
\mathcal{L}_{\text{ctr}}^{(j)} =
- \log \frac{
    \exp(\mathrm{sim}(e_{\text{dist}}^{(j)}, e_{\text{clean}}) / \tau)
}{
    \sum_{k \neq j}  w_{jk} \cdot \exp(\mathrm{sim}(e_{\text{dist}}^{(j)}, e_{\text{dist}}^{(k)}) / \tau)
    + \sum_{l \in \mathcal{B}_{\text{other}}} \exp(\mathrm{sim}(e_{\text{dist}}^{(j)}, e_{\text{other}}^{(l)}) / \tau)
}.
\]

for $\tau = 0.10.$ We use $\mathcal{B} = 256$ clean images per batch, each with $\mathcal{B}_{\text{other}} = 256$ randomly sampled degraded variants from our library of primitive and compound distortions.

The contrastive objective encourages degraded embeddings to reflect distortion structure, but it does not explicitly prevent the clean embedding from drifting toward degradation-sensitive features. We introduce a quality-aware regularizer that penalizes the clean embedding for exhibiting distortion evidence
\[
\mathcal{L}_{\text{qual}}^{(j)} = 
\sum_{c \in d^{(j)}} 
\hat{p}(c \mid e_{\text{clean}}),
\]
where $\hat{p}(c \mid e_{\text{clean}})$ is the predicted probability of distortion $c$ from $e_{\text{clean}}$. This term discourages degradation hallucination and anchors the representation to remain distortion-free for clean inputs.

The final encoder loss is $\mathcal{L}_{\text{CLIP}} = \frac{1}{m}\sum_{j=1}^{m} ( \mathcal{L}_{\text{ctr}}^{(j)} + \mathcal{L}_{\text{qual}}^{(j)})$, yielding embeddings that: (i) align degraded and clean images, (ii) reflect compositional overlap, and (iii) support controllable restoration.

\textbf{Stage 2: Restoring with Conditional Diffusion.}
We adopt a latent diffusion model, Stable Diffusion v1.5 \citep{rombach2022high}. During training, we replace the standard CLIP image encoder with our fine-tuned compound distortion-aware encoder, providing semantic and degradation context. Then, we concatenate the conditioning vector from the image encoder with the text embedding of the restoration prompt from the frozen CLIP text encoder. These embeddings are passed through a learnable cross-attention layer at each UNet block following the original Stable Diffusion v1.5 design for prompt conditioning. This allows PRISM to attend to the desired set of degradations during each denoising step. Unlike sequential approaches \citep{jiang2024autodir}, PRISM conditions on the full composite prompt in a single denoising trajectory, enabling joint restoration of overlapping artifacts. While MPerceiver \citep{ai2024multimodal} encodes multiple degradations as concatenated tokens, this strategy represents multi-distortion inputs but does not explicitly enforce disentangled or compositional structure in the embedding space that supports controlled recombination.

Latent diffusion can blur fine-scale structures critical in scientific imagery. Following \citet{jiang2024autodir}, we integrate a Semantic Content Preservation Module (SCPM), a lightweight decoder-side refinement block that adaptively fuses encoder and decoder features to preserve edges and small textures. Full architectural details and ablations are provided in Appendix~\ref{sec:ablations_supp}.

For fair comparison, all baselines are trained on the fixed set of primitive distortions. Training details and compute requirements are provided in Appendix \ref{sec:training details}, baselines are described in Appendix \ref{sec:baselines}, and ablations over model components are provided in Appendix \ref{sec:ablations_supp}. 

\subsection{Prompting}

PRISM enables both \textit{expert-guided} and \textit{automated} restoration by grounding free-form natural language prompts in a structured task space. 

At inference time, PRISM supports:
\begin{itemize}
    \item \textbf{Expert-guided restoration:} A free-form instruction (e.g., ``remove fog and color shift'') is embedded with the frozen CLIP text encoder and used to condition diffusion.

    \item \textbf{Automated restoration:} Given an input image, a lightweight MLP predicts a multi-label distortion set from the image embedding. This set is converted into a standardized prompt (``remove distortions $x,y,z$'') and encoded by CLIP. 
\end{itemize}

In both prompting scenarios, the text embeddings condition the latent diffusion model through cross-attention at each denoising step. Sensitivity to prompt variation is analyzed in Appendix~\ref{sec:ablations_supp}.

\subsection{Evaluation}

We evaluate PRISM on: (1) compound and controllable restoration, (2) handling unseen real-world composites, and (3) downstream utility. Unless noted otherwise, we evaluate manual restoration with predefined distortion types, using the free-form prompts generated by GPT-4. Full details on datasets and evaluation are in Appendix \ref{sec:dataset} and \ref{sec:evaluation}.

\paragraph{Mixed Degradations Benchmark (MDB).}
We use the held-out subset of the triplets $(I_{\text{clean}}, I_{\text{dist}}, p)$ from our dataset as our fixed testbed. This MDB measures sequential vs. composite prompting and prompt faithfulness under compound degradations. This dataset extends beyond the CDD-11 dataset proposed by \citet{guo2024onerestore} to span a broad set of real-world degradations, with varied intensities.

\paragraph{Handling Unseen Distortions.}
For zero-shot tests, we evaluate on real domains with compound distortions not explicitly seen in training: underwater effects in UIEB \citep{uieb}, under-display camera artifacts \citep{udc}, and fluid distortions \citep{thapa2020dynamic}. These datasets probe PRISM’s ability to extend to novel distortions.

\paragraph{Downstream Utility.}
Standard benchmarks measure pixel similarity to a clean reference, but such metrics fail to capture whether restored images remain scientifically useful. We instead evaluate restoration through downstream tasks using real datasets with natural distortions and undistorted views as ground truth. To reflect how restoration outputs are typically used in practice, we use {off-the-shelf pretrained models}, giving a conservative but practical measure of utility. We test across four domains:
\begin{enumerate}
    \item \textbf{Remote sensing with Sen12MS  \citep{schmitt2019sen12ms}:} landcover classification \citep{papoutsis2023benchmarking} on cloudy satellite data, with labels from cloudless samples.
    \item \textbf{Wildlife monitoring with iWildCam 2022 \citep{beery2022iwildcam}:} species classification with SpeciesNet \citep{gadot2024crop} on low-confidence nighttime images with expert labels from high-confidence frames of the same sequence.
    \item \textbf{Segmentation and fluorescence in microscopy with BioSR \citep{gong2021deep}:} segmentation and fluorescence/intensity measurements of clathrin-coated pits from low signal-to-noise data using MicroSAM \citep{archit2025segment}, compared to high quality structured illumination microscopy (SIM) ground truth.
    \item \textbf{Urban forest monitoring using our novel Rooftop Cityscapes dataset:} panoptic segmentation \citep{lin2017refinenet} of cityscapes under haze/low light, with paired, labeled clear-weather data. See Appendix \ref{sec:evaluation} for details on this custom dataset.
\end{enumerate}
\section{Results and Discussion}
\label{sec:results}

\subsection{Breaking the Cascade: Compound Restoration Made Robust}

\begin{wraptable}{l}{8cm}
    \vspace{-5mm}
    \centering
    \caption{PRISM outperforms baselines with manual prompting on MDB, where each test image has up to three distortions. Best results are \textbf{bolded}, second-best are \underline{underlined}.}
    \scriptsize
    \label{tab:mixed}
    \setlength{\tabcolsep}{5.0pt}
    \begin{tabular}{l|l|c c c c}
        \toprule
        \textbf{Category} & \textbf{Method} & \textbf{PSNR $\uparrow$} & \textbf{SSIM $\uparrow$} & \textbf{FID $\downarrow$} & \textbf{LPIPS $\downarrow$} \\
        \midrule
        \multirow{4}{*}{All-in-One} 
        & AirNet      & 14.76 & 0.742 & 78.55 & 0.382 \\
        & Restormer$_A$ & 16.32 & 0.768 & 70.11 & 0.365 \\
        & NAFNet$_A$ & 16.98 & 0.776 & 68.30 & 0.352 \\
        & PromptIR   & 18.11 & 0.801 & 62.78 & 0.298 \\
        \midrule
        \multirow{3}{*}{Diffusion} 
        & DiffPlugin & 19.07 & 0.821 & 53.88 & 0.255 \\
        & MPerceiver & \underline{20.84} & 0.829 & \textbf{48.18} & \underline{0.235} \\
        & AutoDIR & 20.42 & \underline{0.833} & 50.75 & 0.246 \\
        \midrule
        \multirow{2}{*}{Composite} 
        & OneRestore & 19.36 & 0.812 & 59.42 & 0.276 \\
        & \textbf{PRISM (ours)} & \textbf{22.08} & \textbf{0.842} & \underline{48.97} & \textbf{0.218} \\
        \bottomrule
    \end{tabular}
    \vspace{-4mm}
\end{wraptable}

Sequentially removing distortions often accumulates errors, artifacts, or inconsistencies, while restoring all distortions jointly yields more stable, high-fidelity results. Our MDB evaluation supports this intuition, with qualitative results in Appendix Figs. \ref{fig:mdb_qualitative} and \ref{fig:attention}.

Table \ref{tab:mixed} highlights a divide between early all-in-one models \citep{airnet, restormer, nafnet, promptir}, which are trained per distortion and generalize poorly to mixtures, recent diffusion approaches \cite{liu2024diff, ai2024multimodal, jiang2024autodir}, and models that explicitly target composite restoration \citep{guo2024onerestore}. While OneRestore is trained on composite datasets like PRISM, it fails to match the perceptual quality of diffusion model outputs; however, these diffusion methods typically remain limited by their reliance on single-distortion or sequential training regimes.

PRISM achieves the best results across both fidelity (PSNR/SSIM) and perceptual metrics (FID/LPIPS), owing to two design choices: (1) \textit{compound-aware supervision} over mixed degradations, and (2) \textit{contrastive disentanglement} of distortions. We study the contributions of each below.

\paragraph{Compound-aware supervision supports restoration under increasingly complex mixtures.}

Training on combinatorial mixtures of degradations (full, partial, and negative restoration) allows simultaneous removal of multiple effects without cascading errors. 
\begin{wrapfigure}{r}{0.6\textwidth}
    \centering
    \vspace{-3mm}
    \includegraphics[width=0.6\textwidth]{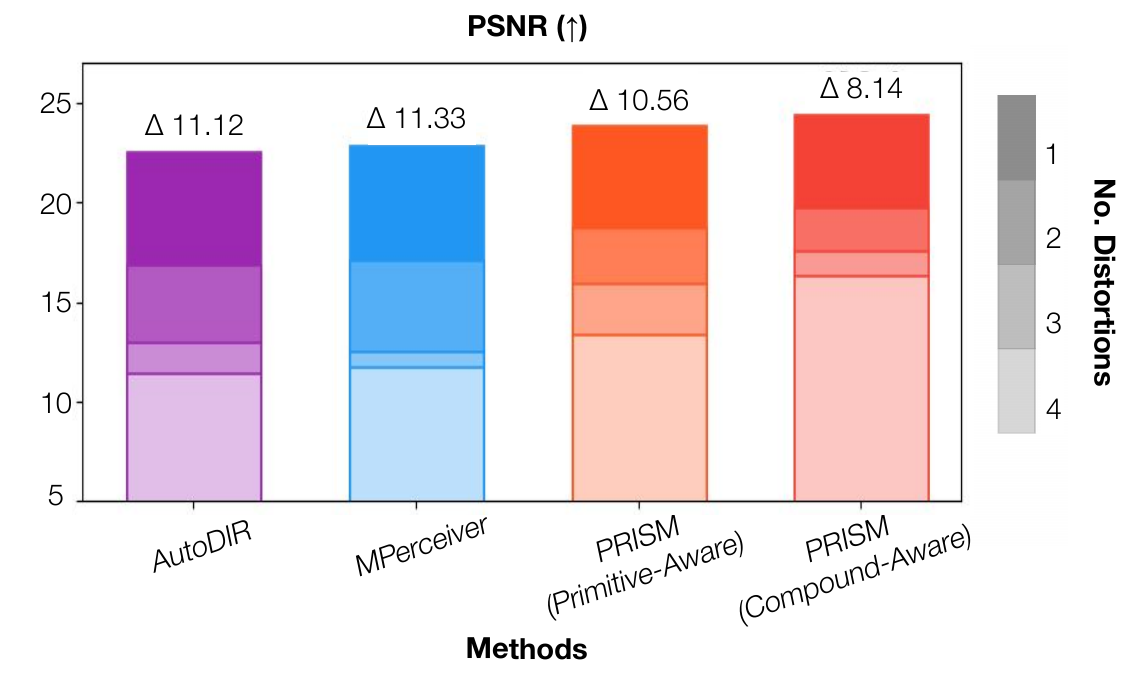}
    \caption{\textit{PRISM trained on composite examples scales best with the number of distortions.} This outperforms our model trained on each degradation separately as well as comparable baselines, emphasized by the $\Delta$ PSNR of test images with 1 vs. 4 distortions) above each bar.}
    \label{fig:composite}
    \vspace{-4mm}
\end{wrapfigure}
Fig. \ref{fig:composite} shows that while PRISM matches baselines on images with a single degradation, it significantly outperforms them as complexity grows, especially on unseen cases with four distortions (an extension of the MDB set).
Training on composites explicitly outperforms training on primitives separately, and that improved image embeddings from our contrastive loss provide an additional boost over baselines. Importantly, this compound-aware structure not only improves robustness under increasing distortion complexity, but also lays the foundation for selective restoration. These trends hold consistently across other metrics (SSIM, LPIPS, and FID), as shown in Appendix Fig. \ref{fig:composite_app}).

\newpage

\paragraph{Contrastive disentanglement improves compositional restoration.} 

\begin{wrapfigure}{l}{0.4\textwidth}
    \centering
    \vspace{-1mm}
    \includegraphics[width=0.4\textwidth]{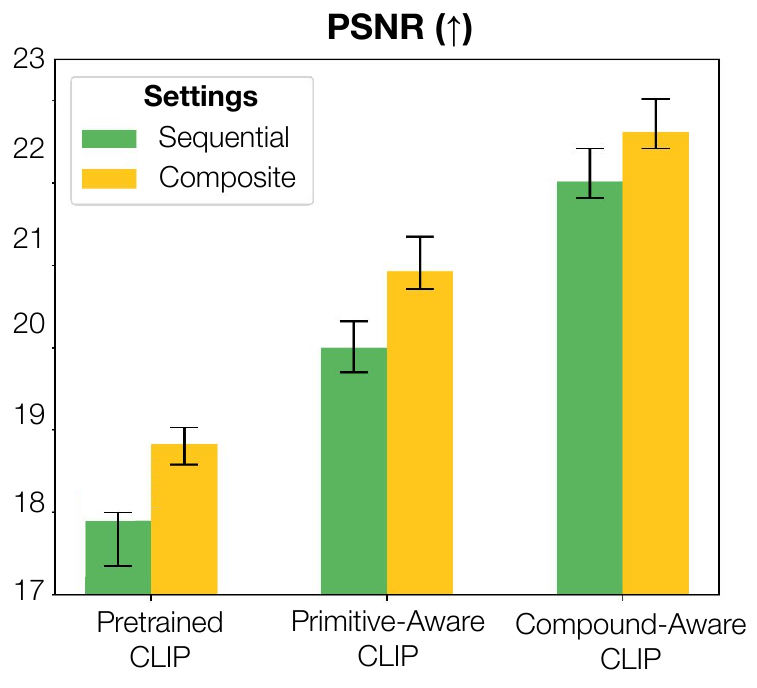}
    \vspace{-6mm}
    \caption{\textit{Latent disentanglement enables both stepwise and single-shot restoration, closing the gap between prompting strategies.}}
    \label{fig:loss_seq}
    \vspace{-5mm}
\end{wrapfigure}

Appendix Fig. \ref{fig:latent} visualizes the effect of the fine-tuning CLIP on the image embedding space. Our compound-aware weighted contrastive loss explicitly incorporates mixed degradations, enforcing a latent geometry that pulls compound distortions toward the span of their primitives (right). This design closes the gap between sequential and single-shot prompting (Fig.~\ref{fig:loss_seq}), ensuring that multi-step restoration behaves predictably and that single-shot prompts achieve comparable fidelity. Appendix Fig.~\ref{fig:loss_app} demonstrates that these improvements generalize beyond PSNR. This allows experts to target specific degradations without disturbing other content. By combining compound-aware supervision with contrastive disentanglement, PRISM not only outperforms existing baselines on restoring compound degradations, but also maintains fine control.

\subsection{Compositionality Enables Adaptive Restoration in Novel Settings}

The compositional structure of the image latent space also supports generalization to unseen degradations. If degradations are represented compositionally, then novel composites can be modeled as combinations of known primitives. This means PRISM can automatically identify constituent distortions and restore them, even if the exact combination was never seen in training.

We evaluate zero-shot restoration on three domains (underwater imagery, under-display cameras, and fluid lensing), each containing complex, previously unseen distortions (Table~\ref{tab:zs_selected}). For each dataset, we use the compound-aware CLIP encoder to identify the fixed set of distortion types present in the images of each dataset. We then apply the same manual prompts over this standardized set for all models to ensure a fair, consistent evaluation. While the predicted distortion categories for UIEB were more variable and often reflected mixtures of multiple effects such as low light, haze, contrast, and color shifts, the POLED and ThapaSet datasets exhibited more uniform and stable classifications, providing clearer mappings to their dominant distortion types: low light, blur, and contrast vs. refraction and warping, respectively. Qualitative examples are shown in Appendix~\ref{sec:add_figures}.

\begin{table*}[h]
    \centering 
    \vspace{-3mm}
    \caption{PRISM achieves state-of-the-art zero-shot performance across underwater (UIEB), under-display camera (POLED), and fluid lensing (ThapaSet) benchmarks. Best results are \textbf{bolded}, second-best are \underline{underlined}.}
    \scriptsize
    \label{tab:zs_selected}
    \setlength{\tabcolsep}{4pt}
    \begin{tabular}{l|l|ccc|ccc|ccc}
        \toprule
        \multirow{2}{*}{\textbf{Category}} &
        \multirow{2}{*}{\textbf{Method}} 
        & \multicolumn{3}{c|}{\textbf{UIEB}} 
        & \multicolumn{3}{c|}{\textbf{POLED}} 
        & \multicolumn{3}{c}{\textbf{ThapaSet}} \\ 
        \cline{3-11}
        & 
        & \rule{0pt}{2.6ex}\textbf{PSNR $\uparrow$} & \textbf{SSIM $\uparrow$} & \textbf{LPIPS $\downarrow$}
        & \textbf{PSNR $\uparrow$} & \textbf{SSIM $\uparrow$} & \textbf{LPIPS $\downarrow$}
        & \textbf{PSNR $\uparrow$} & \textbf{SSIM $\uparrow$} & \textbf{LPIPS $\downarrow$} \\
        \midrule
        \multirow{4}{*}{All-in-One} 
        & AirNet        
            & 17.51 & 0.768 & 0.468 
            & 13.61 & 0.582 & 0.529 
            & 18.41 & 0.388 & 0.609 \\
        & Restormer$_A$ 
            & 18.13 & 0.792 & 0.454 
            & 14.38 & 0.608 & 0.512 
            & 19.31 & 0.413 & 0.588 \\
        & NAFNet$_A$    
            & 17.76 & 0.756 & 0.479 
            & 13.02 & 0.591 & 0.543 
            & 18.94 & 0.401 & 0.596 \\ 
        & PromptIR      
            & 19.76 & 0.858 & 0.417 
            & 16.42 & 0.651 & 0.468 
            & 20.63 & 0.439 & 0.564 \\
        \midrule
        \multirow{3}{*}{Diffusion} 
        & DiffPlugin    
            & 20.72 & 0.874 & 0.392 
            & 17.01 & 0.659 & 0.451 
            & 21.15 & 0.454 & 0.536 \\
        & MPerceiver    
            & \underline{21.18} & \underline{0.889} & \underline{0.366} 
            & \underline{17.55} & \underline{0.669} & 0.436 
            & 21.41 & 0.459 & \underline{0.522} \\
        & AutoDIR       
            & 21.02 & \underline{0.887} & 0.374 
            & 17.33 & 0.665 & \textbf{0.431} 
            & \underline{21.53} & \underline{0.462} & 0.528 \\
        \midrule
        \multirow{2}{*}{Composite} 
        & OneRestore    
            & 20.53 & 0.869 & 0.404 
            & 16.82 & 0.654 & 0.457 
            & 20.82 & 0.446 & 0.548 \\
        & \textbf{PRISM (ours)} 
            & \textbf{22.18} & \textbf{0.914} & \textbf{0.331} 
            & \textbf{18.26} & \textbf{0.694} & \underline{0.419} 
            & \textbf{22.36} & \textbf{0.487} & \textbf{0.492} \\
        \bottomrule
    \end{tabular}
    \vspace{-2mm}
\end{table*}

Both all-in-one models like AirNet and Restormer and composite methods such as OneRestore fall short in capturing complex degradation interactions without robust priors. In contrast, diffusion models like AutoDIR generalize well but often treat distortions in isolation, which may introduce unwanted intermediate errors in linearly restoring inputs.

PRISM’s state-of-the-art zero-shot performance is driven by a latent space that treats compound degradations as structured combinations of primitives. Rather than attempting to model the complex, non-linear physics of every possible distortion interaction, the contrastive design enforces a compositional logic. By training on submixtures, the model learns to represent a composite (e.g., haze + overexposure) as a joint embedding of its constituent parts. This allows the model to map previously unseen mixtures into a known coordinate system defined by established primitives, effectively ``interpolating'' a restoration strategy for novel compounds. Consequently, PRISM avoids the pitfalls of brittle category memorization, enabling the restoration of unseen mixtures using the semantic proximity between known primitives and unknown composites without additional supervision.

\begin{figure}[h]
    \vspace{-3mm}
    \centering
    \includegraphics[width=\textwidth]{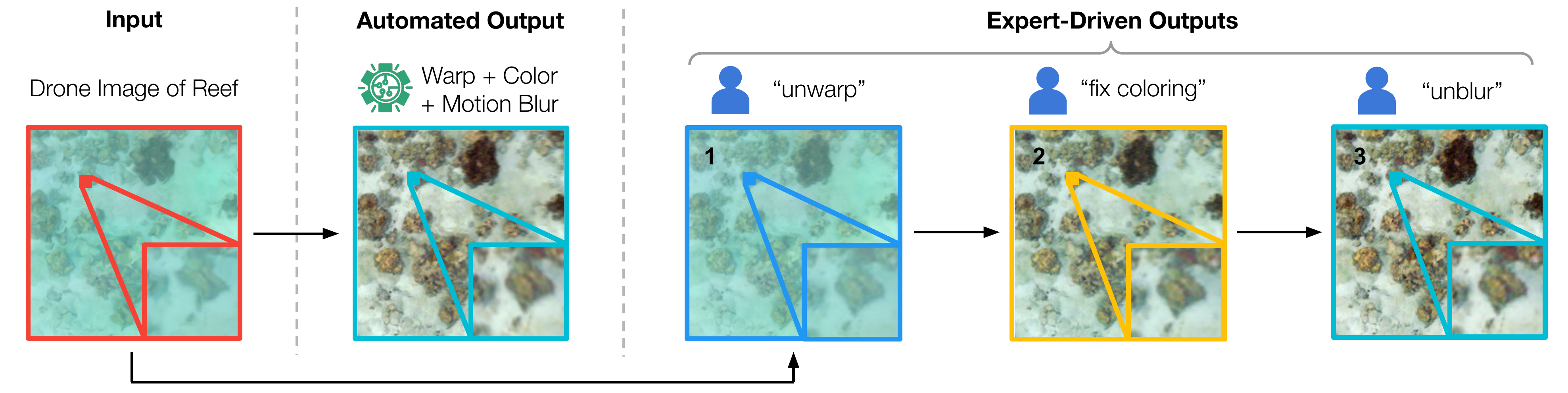}
    \caption{\textit{Structured, compositional latent geometry supports both automated (left) and expert-driven (right) generalization.} With an expert-in-the-loop, prompts progressively target distortions.}
    \label{fig:unseen_latent}
    \vspace{-2mm}
\end{figure}

Fig. \ref{fig:unseen_latent} illustrates how this design translates into practice with high-resolution drone imagery over coral reefs in Mo'orea \citep{saccomanno2025}. While our method enables automated classification and correction of novel distortions, it also provides interpretability and control: experts can explore how distortions relate, refine restoration strategies, and avoid black-box corrections. In improving performance on partial restoration, PRISM enables experts to issue stepwise prompts (e.g., ``unwarp,'' ``fix coloring,'' ``unblur'') to iteratively restore unseen distortions as they see fit.

\subsubsection{The Case for Controllability in Science}
While PRISM restores images robustly under compound degradations, we argue that in some applications, restoration quality cannot be judged by appearance alone. We assess PRISM’s impact on downstream tasks across four domains: remote sensing, ecology, microscopy, and urban monitoring.

This raises a critical question: if a model can remove all degradations, should it? In scientific imaging, often \textit{no}. Distortions can be mixed with faint but meaningful signals, and indiscriminate restoration may erase these cues or introduce artifacts that mislead downstream models. Maintaining downstream fidelity requires control: experts must choose what to correct versus preserve.

As shown in Table~\ref{tab:controllability}, controllability significantly improves downstream performance over full restoration (automatically detecting all distortions present) in three of four domains. For instance, in nighttime camera trap data, restoring only contrast may improve recognition over full restoration, which can blur subtle texture cues. In urban scenes, removing haze improves segmentation, but also brightening the image may over-adjust vegetation in the distance. Remote sensing is the exception: full, automatic restoration performs slightly better, as removing only clouds leaves images under-illuminated and hazy.  

\begin{table}[h]
\vspace{-3mm}
\centering
\scriptsize
\caption{Selective controllability outperforms full restoration across three of four downstream tasks. We report mean $\pm$ std over 3 random seeds. Best results are \textbf{bolded}.}
\begin{tabular}{l|cccc}
\toprule
\textbf{Domain} & \textbf{Degraded Input} & \textbf{Full Restoration} & \textbf{Selective Restoration} & \textbf{p-value} \\
\midrule
Remote sensing (Acc. $\uparrow$) & 0.781 $\pm$ 0.013 & \textbf{0.842 $\pm$ 0.011} & 0.836 $\pm$ 0.012 & 0.11 (n.s.) \\
Camera Traps (Acc. $\uparrow$) & 0.921 $\pm$ 0.004 & 0.976 $\pm$ 0.008 & \textbf{0.984 $\pm$ 0.004} & 0.032 $<$ 0.05 \\
Microscopy (mIoU $\uparrow$)  & 0.353 $\pm$ 0.015 & 0.475 $\pm$ 0.012 & \textbf{0.580 $\pm$ 0.010} & 0.018 $<$ 0.05 \\
Urban scenes (mIoU $\uparrow$) & 0.548 $\pm$ 0.018 & 0.615 $\pm$ 0.014 & \textbf{0.650 $\pm$ 0.012} & 0.041 $<$ 0.05 \\
\bottomrule
\end{tabular}
\label{tab:controllability}
\end{table}

\begin{figure}[h] 
    \vspace{-3mm}
    \centering 
    \includegraphics[width=.9\textwidth]{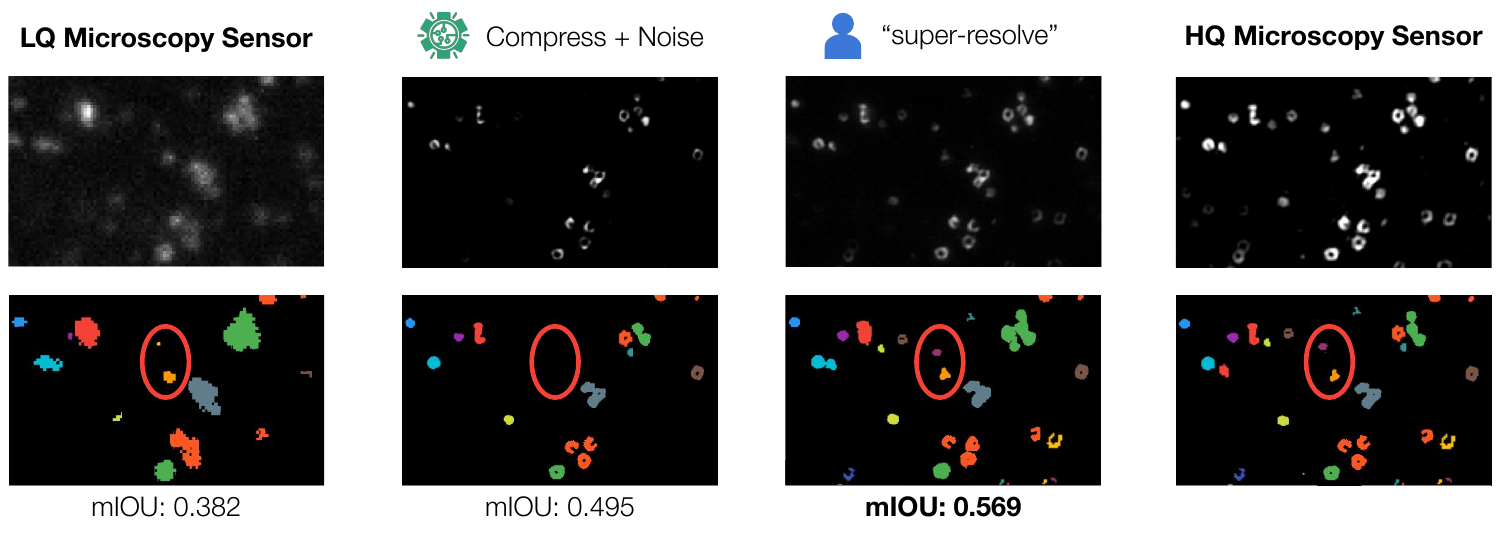} 
    \caption{\textit{Selective restoration improves segmentation of clathrin-coated pits in microscopy.} Super-resolution alone improves mIoU, while automatically detecting and removing noise suppresses faint but biologically relevant signals (see regions encircled in red), reducing accuracy. } 
    \label{fig:microscopy} 
    \vspace{-6mm}
\end{figure}

Fig. \ref{fig:microscopy} illustrates a use case of controllability in microscopy: super-resolution alone improves segmentation alignment with SIM ground truth, but additional denoising erases faint, biologically relevant structures. Additional examples across other domains are included in Appendix \ref{sec:downstream_app}.  

In some cases, \textit{controllability enables different analyses on the same data} which may depend on fundamentally different visual cues. Within microscopy, segmentation of clathrin-coated pits relies on high-frequency structural detail, such as sharp boundaries and local contrast that delineate subcellular objects. In contrast, fluorescence measures the mean pixel intensity within biological regions, a proxy for molecular concentration or protein recruitment. This calculation is highly sensitive to {intensity bias and noise variance} but relatively tolerant to mild blur.

Table~\ref{tab:task_dependent_biosr} summarizes performance across restoration settings. {Super-resolution} yields the best segmentation mIoU by enhancing edges and subcellular boundaries, but this same sharpening increases fluorescence error. {Denoising} has the opposite effect. While it preserves the intensity distribution of input data, producing the lowest fluorescence MSE, it removes fine structure needed for accurate segmentation. Combined restoration produces intermediate or degraded results, demonstrating that no single transformation can satisfy both scientific objectives simultaneously.

\begin{wraptable}{r}{0.45\textwidth}
\centering
\scriptsize
\vspace{-5mm}
\caption{\textit{Task-dependent restoration performance on BioSR.} Segmentation prefers structural enhancement, whereas fluorescence quantification requires accurate intensity preservation. No single restoration setting is optimal for both tasks.}
\label{tab:task_dependent_biosr}
\begin{tabular}{lcc}
\toprule
\textbf{Restoration Prompt} 
    & \textbf{Segmentation} 
    & \textbf{Fluorescence} \\
\textbf{} 
    & \textbf{mIoU $\uparrow$} 
    & \textbf{MSE $\downarrow$} \\
\midrule
Denoise          & 0.479 & \textbf{0.006
} \\
Super-Resolution & \textbf{0.580} & 0.018 \\
Combined         & 0.475 & 0.014 \\
No restoration   & 0.356 & 0.025 \\
\bottomrule
\end{tabular}
\vspace{-3mm}
\end{wraptable}

These findings highlight that \textit{restoration is task-dependent}, and experts may not always know how to preprocess their data beforehand. Blanket restoration can overlook the varying demands of scientific workflows. PRISM’s controllability enables experts to ensure that restored images remain faithful to the underlying scientific signal as needed.

Important challenges remain for PRISM. Our training still depends on synthetic augmentations that cannot fully capture real distortions. Moreover, extending controllability beyond specifying which distortions to remove to their intensity and spatial extent would enable localized restoration and finer-grained preservation of scientific signals. 

While diffusion-based restoration typically demands greater computational resources than traditional encoder–decoder architectures, PRISM maintains competitive runtimes. As detailed in Appendix~\ref{sec:ablations_supp} (Table~\ref{tab:latency_cost}), its latency is comparable to existing state-of-the-art diffusion frameworks. In this way, PRISM balances practical efficiency with addressing key priorities in scientific restoration.
\vspace{-3mm}
\section{Conclusions}
\vspace{-3mm}

Our results show that controllable \textit{and} compound-aware restoration is critical for scientific and environmental imaging. PRISM outperforms both specialized and generalist baselines by combining (1) compound-aware supervision, which exposes the model to overlapping degradations, and (2) weighted contrastive disentanglement, which organizes the latent space according to composite distortions. Together, these yield more robust and interpretable restoration.

We also find strong generalization beyond curated training sets. PRISM achieves robust zero-shot performance on underwater imaging, under-display camera correction, and fluid lensing, showing that compositional representations extend naturally to unseen domains. Importantly, evaluations on real composite degradations confirm generalization beyond our synthetic training pipeline.  

A key insight is that \emph{more restoration is not always better}. Across diverse domains, we show that indiscriminate removal of degradations suppresses faint but meaningful signals or introduces artifacts. Moreover, the appropriate level and type of restoration can be \emph{task-dependent}, as different scientific analyses may rely on distinct visual cues and therefore benefit from different restoration strategies. Allowing experts to choose which distortions to correct is essential for scientific precision. 

\newpage
\clearpage

\subsubsection*{Acknowledgments}
This work was supported in part by a Schmidt Science's AI2050 Early Career Fellowship, NSF CAREER Grant (Award No. 2441060), an NSF Graduate Research Fellowship (Award No. DGE-2141064), the NSF and NSERC AI and Biodiversity Change Global Center (NSF Award No. 2330423 and NSERC Award No. 585136), and the MIT Generative AI Consortium. 

\subsubsection*{Ethics Statement}

This work relies exclusively on publicly available datasets and synthetic distortions; all source data are cited appropriately. We acknowledge that restoration models can be misapplied or introduce misleading artifacts, and note that our system is not intended for high-stakes decision-making without domain expert oversight, but as an overall framework design for handling complex real-world data.

\subsubsection*{Reproducibility Statement}

Our \href{https://github.com/RupaKurinchiVendhan/PRISM}{code repository} includes reproducible pipelines for data generation, model training, inference, and evaluation across standard and downstream benchmarks. Detailed descriptions of the dataset preprocessing, model architecture, hyperparameters, training objectives, and evaluation protocols are included in the main paper and appendix, along with extensive ablation studies and full experimental results. All baselines use publicly available implementations following the original authors’ specifications, and we additionally release our benchmarking dataset for mixed degradation removal and evaluation protocols for all downstream task analysis.

\bibliographystyle{iclr2025_conference}
\bibliography{iclr2025_conference}

\newpage
\appendix
\section{Training Details}
\label{sec:training details}

PRISM is trained on 8$\times$40GB NVIDIA A100 GPUs using mixed-precision training (FP16).

\paragraph{Hyperparameters.}  
For our final model we use:
\begin{itemize}
    \item Batch size: 264 (global, distributed across 8 GPUs)  
    \item Learning rate: $1\times 10^{-4}$ (AdamW optimizer)  
    \item No learning-rate warm-up; cosine decay schedule  
    \item Training epochs: 500  
    \item Input resolution: $256\times256$  
    \item Gradient clipping: 1.0  
    \item EMA decay: 0.9999 for model weights  
\end{itemize}

We initialize the backbone from publicly available Stable Diffusion v1.5 weights \citep{rombach2022high}.  

\paragraph{CLIP Fine-Tuning.}  
For the embedding space, we initialize from OpenAI CLIP ViT-B/32 \citep{radford2021learning} pretrained weights. We fine-tune only the final projection layers and cross-attention adapters, freezing the base vision and text encoders to preserve semantic alignment. Fine-tuning uses:
\begin{itemize}
    \item Batch size: 256  
    \item Learning rate: $5\times 10^{-5}$ (AdamW optimizer)  
    \item Training epochs: 50   
\end{itemize}

\paragraph{Distortion Classifier for Automated Restoration.}  
We train the distortion classifier after fine-tuning CLIP, 
using the frozen distortion-aware CLIP image encoder as a feature extractor. The two-layer MLP has a hidden dimension of 512 and predicts over a fixed vocabulary of 
$K = 14$ primitive distortions (blur, haze, low light, color shift, rain, snow, noise, etc.). 

We train the classifier using the same synthetic mixed-degradation dataset used to fine-tune the CLIP image encoder,
treating each degraded image $I$ as a multi-label instance with ground-truth distortion set
$\mathcal{D} \subseteq \{1,\dots,K\}$. 
The label set is converted into a binary target vector 
$\mathbf{y} \in \{0,1\}^K$, and we optimize a binary cross-entropy loss.
We use parameters
\begin{itemize}
    \item Batch size: 512
    \item Learning rate: $1\times 10^{-4}$ (AdamW optimizer)
    \item Training epochs: 98
    \item Cosine learning-rate decay schedule
\end{itemize}
The CLIP image encoder is kept frozen during this stage; only the MLP parameters are updated.
At test time, we predict the distortion set as 
$\hat{d} = \{\, k : \hat{p}_k > 0.85 \,\}$ and convert $\hat{d}$
into the associated standardized prompt ``remove the effects of distortions $d_1, d_2, \dots, d_m$''.

\section{MDB Construction}
\label{sec:dataset}

As stated in Section \ref{sec:methods}, our composite degradation dataset used for ground truth during training was drawn from a diverse collection of datasets spanning multiple scientific and environmental imaging domains. Table \ref{tab:dataset_summary} summarizes the datasets used.
\label{sec:data}
\begin{table}[h]
\centering
\scriptsize
\caption{Summary of Training Datasets. PRISM is trained on a diverse set of natural and scientific domains spanning ecological, medical, astronomical, and remote sensing imagery.}
\label{tab:dataset_summary}
\begin{tabular}{l p{6.5cm} c}
\toprule
\textbf{Dataset} & \textbf{Description} & \textbf{Size} \\
\midrule
\textbf{ImageNet} \citep{deng2009imagenet} & 1000-class benchmark of natural images with visually diverse scenes. & 1.2M \\
\midrule
\textbf{Sen12MS (Sentinel-2)} \citep{schmitt2019sen12ms} & RGB satellite image patches with and without clouds, used for land-cover and cloud-removal tasks. & 720K \\
\midrule
\textbf{iWildCam 2022} \citep{beery2022iwildcam} & Camera trap sequences for wildlife monitoring under challenging lighting and environmental conditions. & 28.8K \\
\midrule
\textbf{EUVP} \citep{islam2020fast} & Paired underwater photos with clear vs. distorted conditions (enhanced clean split). & 3.7K \\
\midrule
\textbf{Cityscapes} \citep{cordts2016cityscapes} & Urban street scenes captured from vehicle-mounted cameras (includes 5K labeled and 20K ``extra''). & 25K \\
\midrule
\textbf{BioSR} \citep{gong2021deep} & Fluorescence microscopy slides (wide-field vs. SIM ground truth) for super-resolution and denoising tasks. & 14K \\
\midrule
\textbf{Brain Tumor MRI} \citep{msoud_nickparvar_2021} & Clinical MRI scans for tumor detection and segmentation with paired clean/noisy modalities. & 7K \\
\midrule
\textbf{AstroSR HSC Surveys} \citep{miao2024astrosr} & Wide-field sky survey images from the Hyper Suprime-Cam (HSC), used for astrophysical source recovery. & 2K \\
\bottomrule
\end{tabular}
\end{table}

Each clean image is transformed into multiple distorted counterparts so that every distortion type is applied across the full dataset. For each image, we apply $N$ distortions. We limit $N$ to a maximum of 3 distortions because we found that $N \geq 4$ degraded the signal significantly and made the task of restoration too difficult (see Appendix E for the relevant ablation over $N$). 

\begin{wrapfigure}{R}{0.5\textwidth}
    \centering
    \vspace{-5mm}
    \includegraphics[width=0.5\textwidth]{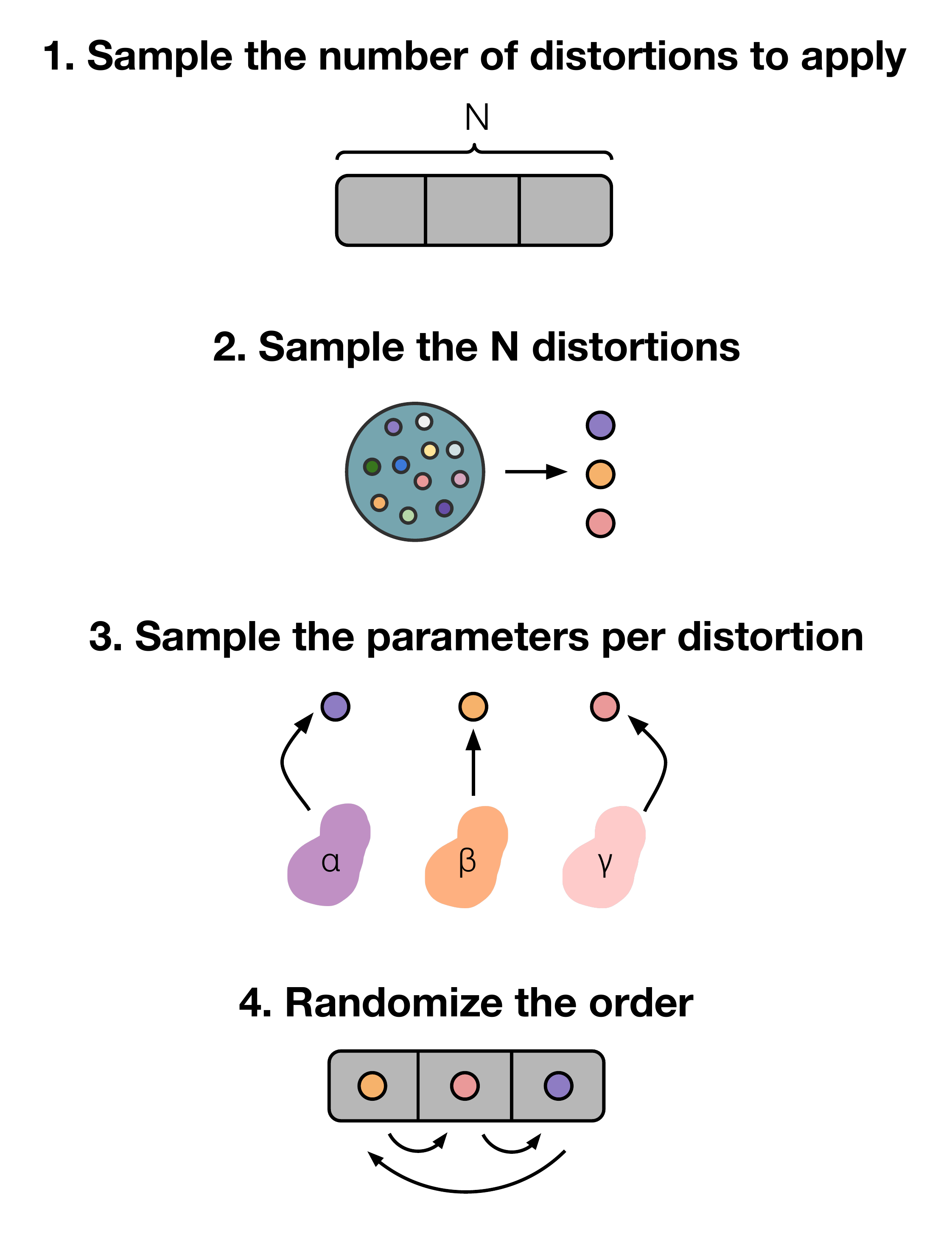}
    \caption{An overview of the data augmentation pipeline of diverse compound degradations.}
    \label{fig:data_aug}
    \vspace{-5mm}
\end{wrapfigure}

Given the sampled number $N$, distinct distortion types are drawn uniformly from a predefined library $\mathcal{D}$ of transformations. Our distortion set spans 14 categories: including geometric distortions (motion blur, warping, refraction, defocus blur), photometric degradations (contrast, color shifts, brightness, low light), occlusions (clouds, haze, rain, snow), and noise-based effects (additive noise, compression). 

\textbf{Geometric Distortions.} \textit{Elastic Deformation} generates random displacement fields from $\mathcal{U}(-1, 1)$, applies Gaussian smoothing ($\sigma \in [20, 30]$), scales by $\alpha \in [10, 20]$, and warps images via bilinear interpolation. \textit{Refraction} uses Gaussian-filtered noise ($\sigma=10$) with strength $\in [20, 80]$ applied to coordinate maps. \textit{Motion Blur} creates directional kernels (size $\in [5, 10]$) along horizontal, vertical, or diagonal axes, with optional depth-aware masking that selectively blurs foreground or background regions based on threshold $\tau \in [0.3, 0.7]$.

\textbf{Photometric Distortions.} \textit{Low Light} applies multiplicative brightness reduction with factor $f \in [0.4, 0.9]$. \textit{Color Jitter} combines independent RGB channel shifts ($\delta \in [-0.4, 0.4]$) with color casts (warm/cool/green/magenta/cyan/yellow) at intensity $\alpha \in [0.1, 0.3]$. \textit{Overexposure} uses inverse gamma correction ($\gamma = 1/f$, $f \in [1.0, 1.5]$), brightness boost, and soft highlight clipping above threshold $\tau \in [0.4, 0.9]$. \textit{Underexposure} applies gamma with $f \in [0.5, 0.9]$, shadow compression below $\tau \in [0.1, 0.3]$, and luminance-weighted noise ($\sigma \in [0.02, 0.08]$) in dark regions. \textit{Contrast} and \textit{Saturation} use random factors from $[0.4, 1.0]$.

\textbf{Occlusion Effects.} \textit{Haze} blends images with white layers using depth-scaled opacity $\alpha \in [0.65, 0.9]$: $I_{haze} = I(1-D\alpha) + D\alpha$. \textit{Rain} generates two-scale streaks (blur kernels $\in \{7\text{-}23\}$, zoom $\in [1.0, 3.5]$) with selective fog (90th percentile depth, visibility $[8000, 15000]$) and opacity $\alpha \in [0.2, 0.4]$. \textit{Snow} uses multi-scale particle generation with screen blending and minimal fog (95th percentile, visibility $[10000, 20000]$). \textit{Clouds} employs Perlin noise with randomized opacity $\in [0.7, 1.0]$, shadow intensity $\in [0.2, 0.7]$, and blur scaling $\in [1.0, 3.0]$. \textit{Raindrops} places 20-60 drops with radii $\in [3, 50]$ pixels and edge darkening $\in [0.4, 0.8]$.

\textbf{Noise and Resolution.} \textit{Gaussian Noise} randomly applies either Gaussian ($\mathcal{N}(0, \sigma)$, $\sigma \in [0.05, 0.1]$) or salt-and-pepper noise (2-8\% corruption, 50:50 ratio) with equal probability. \textit{Defocus Blur} uses Gaussian kernels with size $k \in [3, 19]$. \textit{Pixelation} downsamples by factors $s \in [2.0, 4.0]$ using bicubic interpolation for super-resolution tasks.

Parameter ranges for each degradation type are uniformly sampled within physically realistic bounds. See our codebase for the full implementation of this distortion library. Finally, the selected distortions are randomly ordered and sequentially applied to the clean image. Randomizing the application order reflects the non-commutative nature of compound distortions and further increases the diversity and realism of the visual outcomes. For all randomized sampling, we used a fixed seed of 42. This data augmentation process is summarized in Fig. \ref{fig:data_aug}.

Prompts describing distortions are auto‑generated with GPT‑4 \citep{hurst2024gpt} to simulate the variability in natural‑language queries likely to be provided at inference time. For each primitive distortion type $c$ in our vocabulary (e.g., haze, blur, noise), we query GPT‑4 with ``Provide 50 different ways a user might ask to remove {\textit{c}} from an image.'' This yields multiple phrasings per distortion (e.g., ``remove haze,'' ``dehaze this image,'' ``clear the atmospheric fog,'' ``fix the hazy artifacts''), promoting robustness to linguistic variation and paraphrasing.

To model realistic multi‑artifact scenarios, we also generate compound prompts. For each pair or triplet of distortions $d = {c_1, c_2, \ldots}$ sampled during data construction, we query GPT‑4 with ``Provide 50 ways a user might ask to remove {c1, c2, ...} from an image.'' GPT‑4 is required to mention all distortions in $d$ within a single natural instruction (e.g., ``remove blur and color shift,'' ``fix the blurriness and correct the color cast,'' ``clean up both the motion blur and the hue distortion''). These compound prompts ensure that the model is trained on natural‑language descriptions of \textit{mixtures} rather than isolated categories.

All GPT‑4 outputs were manually inspected to remove malformed or incomplete phrasings (e.g., missing distortion names, ambiguous requests). Our MDB dataset release includes our image pairs and natural-language prompts, as well as scripts used to generate and sample prompts.

To improve controllability, we further incorporate:
\begin{itemize}
    \item \textbf{Partial prompts}: instructing the model to remove only a subset of degradations present in $I_{\text{dist}}$, requiring the model to learn selective restoration (e.g., input degraded with haze + rain + blur, prompt: ``remove haze and blur'').  
    \item \textbf{Negative prompts}: instructing the model to remove degradations that are \emph{absent}, which enforces that restoration actions are conditional on both image evidence and textual prompts. For instance, if the input is degraded with haze + blur and the prompt is ``remove snow,'' the model should leave the image unchanged with respect to snow.  
\end{itemize}
Approximately 20\% of the training samples are partial prompts and 10\% are negative prompts. The inclusion of partial and negative prompts is critical for teaching PRISM to respect expert instructions. Without them, the model tends to over-restore, indiscriminately removing all degradations it detects. By explicitly training on examples where the correct action is \emph{not} to remove a present or absent degradation, PRISM learns to balance restoration fidelity with adherence to prompts.  

\begin{figure}
    \centering
    \includegraphics[width=\linewidth]{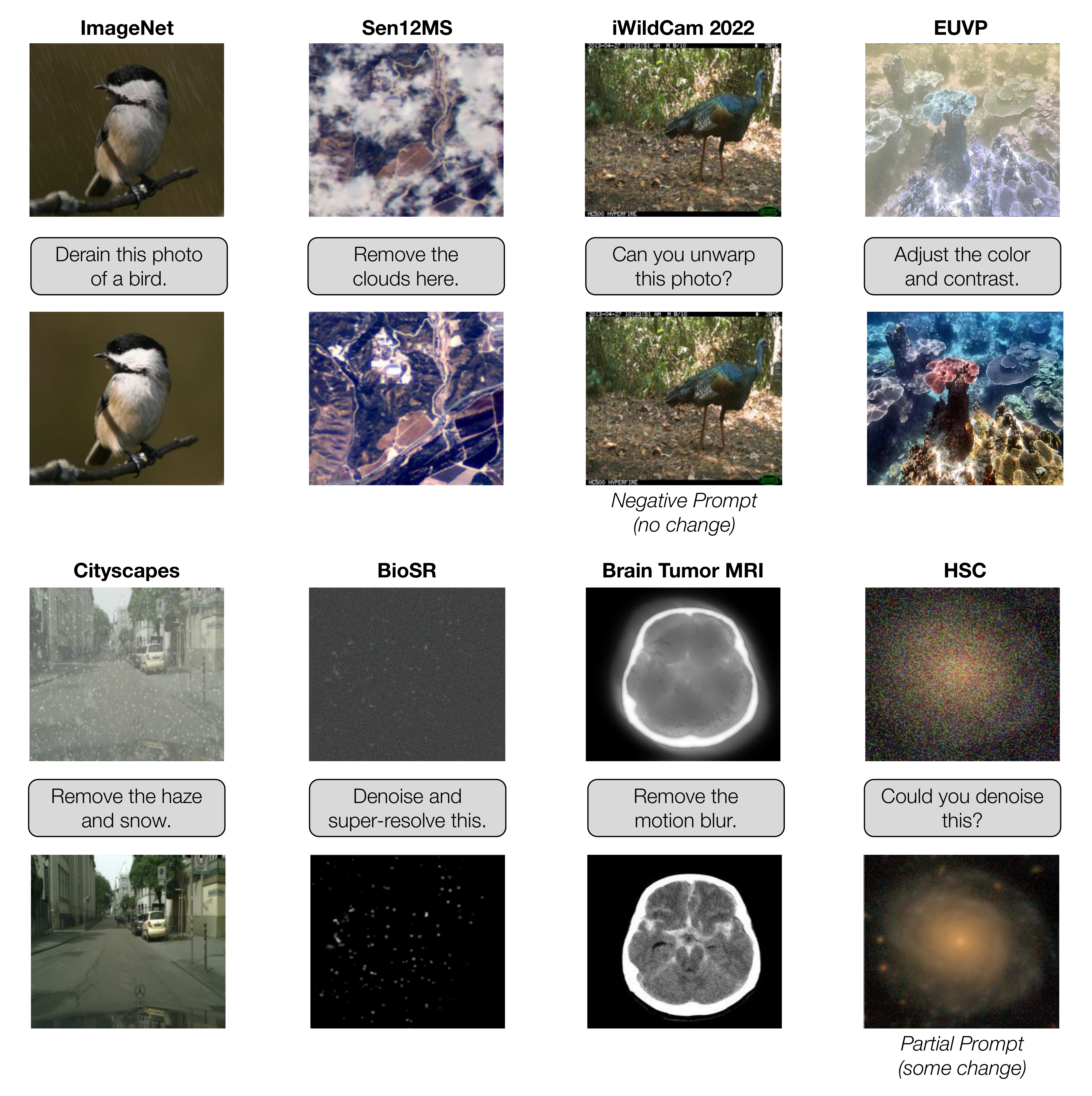}
    \caption{\textit{MDB Examples.} Samples from our compound degradation dataset, with paired ``clean'' (top) and ``distorted'' (bottom) images, with corresponding prompts in between.}
    \label{fig:data_examples}
\end{figure}

The final training corpus consists of approximately $2.5$M triplets, split $80/19/1$ across training, validation, and held-out test sets. For each clean image, multiple degraded variants and prompts are generated, increasing coverage of both degradation mixtures and linguistic variability.  Fig. \ref{fig:data_examples} demonstrates our dataset diversity, with examples of clean, distorted, and prompt triplets.

\section{Evaluation}
\label{sec:evaluation}

As discussed in Section \ref{sec:methods}, we evaluate PRISM along three complementary axes: (1) restoration under synthetic compound degradations, (2) downstream utility in real scientific datasets, and (3) zero-shot robustness to unseen real-world distortions. A summary of each evaluation testbed is provided in Table \ref{tab:evaluation_summary}. Unless noted otherwise, all outputs are generated using a fixed random seed 42.

\begin{table}[h]
\centering
\caption{We group evaluations into (1) synthetic compound degradations, (2) downstream utility in real scientific datasets, and (3) zero-shot robustness to unseen real-world distortions.}
\label{tab:evaluation_summary}
\scriptsize
\begin{tabular}{l|p{4.5cm}|c|p{3.2cm}}
\toprule
\textbf{Evaluation Setting} & \textbf{Task / Dataset} & \textbf{Dataset Size} & \textbf{Metric} \\
\midrule
\multicolumn{4}{l}{\textbf{(1) Synthetic Compound Degradations}} \\
\midrule
MDB & Synthetic mixtures from clean datasets & 25K & PSNR, SSIM, FID, LPIPS \\
\midrule
\multicolumn{4}{l}{\textbf{(3) Zero-Shot Robustness to Unseen Real-World Distortions}} \\
\midrule
Underwater Imaging & UIEB \citep{uieb} & 890 & PSNR, SSIM, LPIPS \\
Under-Display Cameras & POLED (UDC) \citep{udc} & 512 & PSNR, SSIM, LPIPS \\
Fluid Lensing & ThapaSet & 600 & PSNR, SSIM, LPIPS \\
\midrule
\multicolumn{4}{l}{\textbf{(2) Downstream Utility in Real Scientific Datasets}} \\
\midrule
Land Cover Classification & Sen12MS-CR (cloudy Sentinel-2 patches)\newline  \citep{ebel2020multisensor, schmitt2019sen12ms} & 200 & Classification accuracy (ResNet50) \\
Wildlife Classification & iWildCam 2022 (camera traps)\newline  \citep{beery2022iwildcam} & 200 & Classification accuracy (SpeciesNet) \\
Microscopy Segmentation & BioSR (WF vs. SIM microscopy)\newline  \citep{gong2021deep} & 10K & Instance segmentation mIoU (MicroSAM) \&\newline Fluorescence quantification (MSE) \\
Urban Scene Understanding & Rooftop Cityscapes & 5K & Panoptic segmentation mIoU (RefineNet) \\
\bottomrule
\end{tabular}
\end{table}

\paragraph{Downstream Utility.} 
Here, we provide specific details about how we constructed our novel benchmarking suite for evaluation over downstream utility. We re-purpose real datasets with distortions that present known challenges for models across remote sensing, wildlife monitoring, microscopy and weather, where ground truth labels are available \textit{not} because the distorted images are annotated, but because undistorted views exist at different points in time or are collected from a more sophisticated sensor.

Rather than training task-specific models for each of these downstream tasks, we deliberately use off-the-shelf pretrained models. This design choice reflects a realistic scenario: domain experts are far more likely to apply widely available models for segmentation than to train bespoke models for each experimental setup. Using off-the-shelf models therefore provides a conservative estimate of restoration utility in practice and avoids confounding performance gains from joint training on dataset-specific distributions. If restoration improves the outputs of a generic model, this strongly suggests practical downstream utility beyond controlled benchmarks. In each of the four domains, we examine restoration performance on a specified distortion against the default set of automatically-detected distortions present in the input image. We do not compare against domain-specific restoration models (e.g., dedicated cloud removal networks) because our goal is to evaluate generalist models that can flexibly handle a wide variety of distortions; this broader applicability makes them more useful in scientific imaging, where degradations are diverse, overlapping, and often domain-shifted.

\begin{figure}[h] 
    \vspace{-3mm}
    \centering 
    \includegraphics[width=\textwidth]{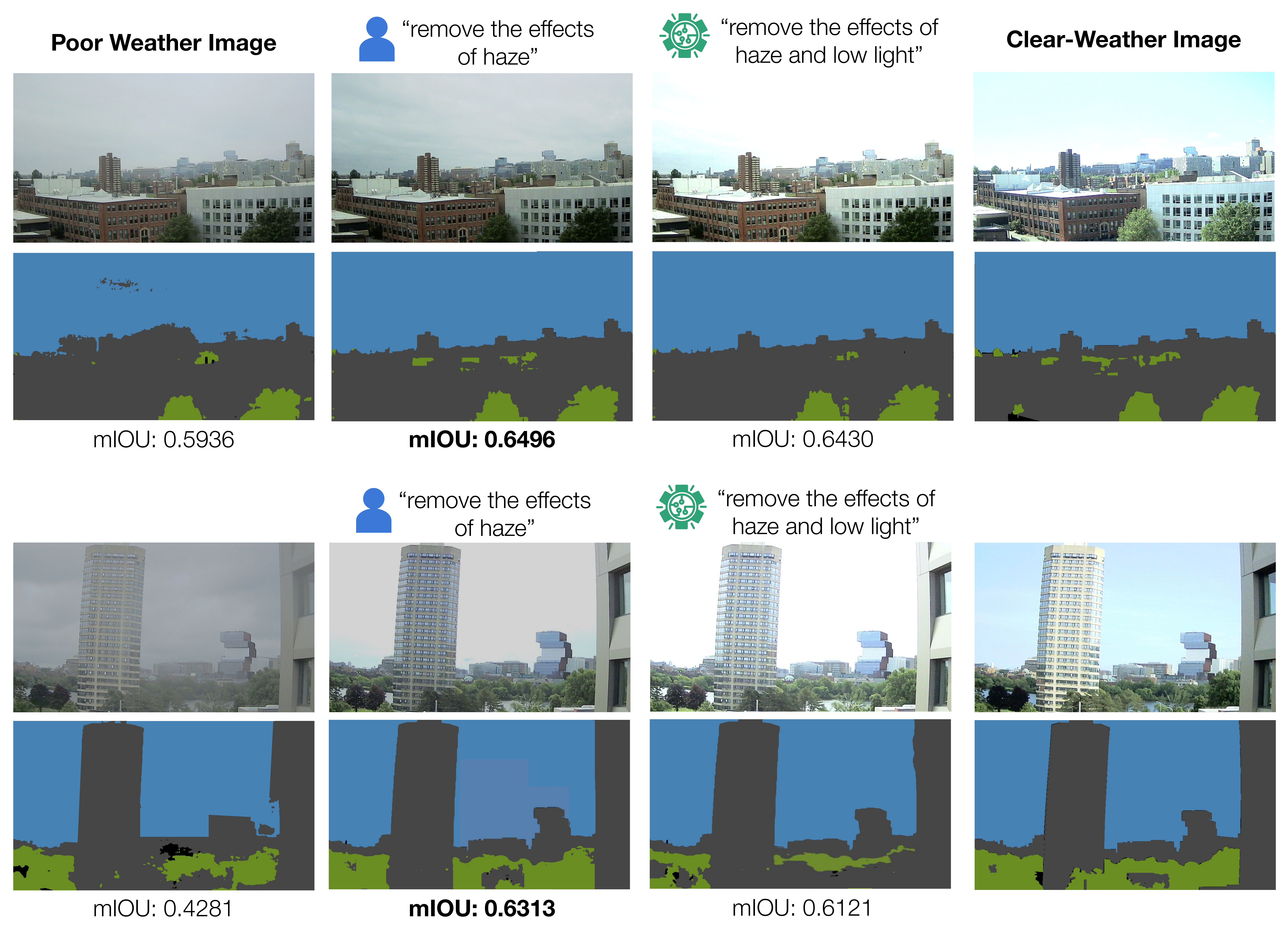} 
    \caption{\textit{Selective restoration helps with urban scene understanding under haze and low light.} Rooftop Cityscapes examples show how selective restoration affects scene understanding. Removing haze alone improves mIoU, while attempting to also remove low light leads to over-correction and lower segmentation accuracy.} 
    \label{fig:rooftops} 
    \vspace{-3mm}
\end{figure}

\begin{enumerate}
    \item \textbf{Land Cover Classification:} To assess performance on land cover classification over satellite data, we select 200 cloudy satellite images degraded by cloud cover from the Sen12MS-CR dataset \citep{ebel2020multisensor}, with land cover labels derived from temporally aligned, cloud-free views in the Sen12MS dataset \citep{schmitt2019sen12ms}. We evaluate using a ResNet50, trained on satellite imagery that includes minimal cloud cover \citep{papoutsis2023benchmarking}. It is important to evaluate on land cover classification because accurate identification of surface types (e.g., forests, croplands, urban areas) under degraded conditions like cloud cover directly underpins large-scale monitoring of climate change, biodiversity, and resource management.
    \item \textbf{Wildlife Classification} Camera trap classification is critical for ecological monitoring, enabling large-scale biodiversity surveys without direct human observation. We evaluate our model on the task of species identification using \textit{iWildCam 2022 Camera Trap} dataset \citep{beery2022iwildcam}. Specifically, we use 200 nighttime wildlife images, after filtering out blank frames (no species) and sample frames with low confidence species predictions (< 0.70). Ground truth annotations are sourced from expert labels of alternate frames in the same camera trap sequence. Classification is evaluated using SpeciesNet \citep{gadot2024crop}, a classifier trained on over 6 million camera trap images.
    \item \textbf{Microscopy Segmentation} Next, we evaluate our model on the task of microscopy image segmentation, which informs the quantification of organelle morphology and dynamics, which are central to understanding cell physiology and disease. We build on the \textit{BioSR dataset} \citep{gong2021deep} introduced by Qiao et al. This dataset was acquired using paired low resolution wide-field (WF, diffraction-limited) and super-resolved structured illumination microscopy (SIM) images of cellular structures (clathrin-coated pits) across a wide range of signal-to-noise ratios. In our setting, the WF images serve as noisy ``distorted'' inputs, while the corresponding high-SNR SIM sensor data provide the undistorted reference. This setup allows us to evaluate restoration not against simulated degradations, but against experimentally aligned ground truth. We measure performance by applying restored images to the downstream task of segmentation, using the microscopy foundation model MicroSAM model \citep{archit2025segment} to generate cell-structure masks, and report segmentation accuracy compared to the high-quality SIM annotations. This mirrors real-world use, where quantitative biological conclusions (e.g., about organelle morphology or cytoskeletal organization) depend critically on reliable segmentation.
    \item \textbf{Urban Scene Understanding} We also evaluate our model on the task of cityscape scene understanding, which enables reliable monitoring of urban forests. To do so, we collected, labeled, and processed a novel \textit{Rooftop Cityscapes} dataset for an additional setting: and haze in urban scenes. Specifically, we deployed fixed-position cameras on several building rooftops across multiple days under varying weather and lighting conditions. From each sequence, we manually identified and labeled frames with clear conditions to serve as the ground truth. We applied an off-the-shelf panoptic segmentation model pretrained on the original Cityscapes \citep{cordts2016cityscapes} dataset to each distorted and restored frame. To ensure reliable comparison, we restricted evaluation to ``stationary'' classes (buildings, vegetation, and sky) while ignoring dynamic objects such as cars or pedestrians, which may change across frames and introduce label inconsistency. See Fig. \ref{fig:rooftops} for qualitative examples from this custom dataset.
\end{enumerate}

\paragraph{Statistical Significance Evaluation}

To assess whether Selective Restoration provided a statistically significant improvement over Full Restoration, we conducted paired hypothesis tests across repeated experimental runs. Each model was trained and evaluated with multiple random seeds on the same dataset splits (seeds 2, 42, and 420), yielding a distribution of results for each condition.

For every domain and downstream task, we computed paired differences between the two methods:

$$
d_i = \text{Selective}_i - \text{Full}_i, \quad i = 1, \dots, n
$$

where $n$ is the number of runs (seeds/splits). This controls for variability due to dataset sampling and ensures that each comparison is made under identical conditions.

We then applied a two-tailed paired t-test to the differences ${d_i}$:

$$
t = \frac{\bar{d}}{s_d / \sqrt{n}}, \quad s_d = \sqrt{\frac{1}{n-1} \sum_{i=1}^n (d_i - \bar{d})^2},
$$

with $n-1$ degrees of freedom. The null hypothesis $H_0$ is that Selective and Full Restoration perform equally ($\mu_d = 0$). The p-value is computed as:

$$
p = 2 \cdot P\!\left(T_{n-1} \geq |t|\right),
$$

where $T_{n-1}$ is a Student’s $t$ distribution with $n-1$ degrees of freedom.
\begin{enumerate}
    \item If $p < 0.05$, we reject $H_0$ and conclude that Selective Restoration significantly outperforms Full Restoration.
    \item If $p \geq 0.05$, we fail to reject $H_0$, indicating that the observed difference may be due to random variation.
\end{enumerate}

\section{Baselines} 
\label{sec:baselines}

AirNet \citep{airnet}, Restormer \citep{restormer}, and NAFNet \citep{nafnet} represent strong encoder–decoder backbones widely used for low-level vision tasks, but they operate in an all-in-one setting without explicit modeling of compound effects. OneRestore \citep{guo2024onerestore} and PromptIR \citep{potlapalli2023promptir} extend this to multi-degradation scenarios: OneRestore introduces a one-to-composite mapping, while PromptIR conditions restoration on learned prompt embeddings. DiffPlugin \citep{liu2024diff} and MPerceiver \citep{ai2024multimodal} adopt modular or token-based conditioning, with DiffPlugin integrating contrastive prompt modules and MPerceiver encoding multiple degradation tokens. AutoDIR \citep{jiang2024autodir} represents a task-routing approach, selecting subtasks adaptively during inference.

Among these, only PRISM employs a weighted contrastive loss to enforce compositional disentanglement in the embedding space. All other baselines use their standard supervision without this contrastive component.

All baseline models were re-trained or fine-tuned on the same set of primitive distortions as PRISM to ensure a fair comparison. This controls for training data bias and isolates differences in architecture and supervision. Following their original training protocol, all baselines (with the exception of OneRestore \citep{guo2019haze}) are trained on single distortions/primitives only, unlike PRISM which is trained on the full combinatorial mixutre set. For evaluation, we predefined a set of primitive degradations (e.g., blur, noise, haze, rain) and applied restoration pipelines consistently across models, so that all methods operated under identical inputs whether they were trained to remove distortoins independently or compositely. This avoids favoring baselines tailored to a specific distribution and provides a controlled setting for compound restoration.

Together, these baselines span backbone, prompt-driven, and diffusion-based strategies. Further details for fine-tuning and re-training our baselines and access to their implementations are included in the provided codebase linked above.

\section{Additional Ablations and Sensitivity Analysis}
\label{sec:ablations_supp}

To better understand the contributions of individual design choices in PRISM, we conduct ablations on our held-out the Mixed Degradations Benchmark (MDB) test set.

\paragraph{Semantic Content Preservation Module (SCPM).} 
While latent diffusion excels at high-level denoising, it often loses fine-grained semantic or structural details critical in scientific imagery. To address this, we integrate the Semantic Content Preservation Module (SCPM), a lightweight decoder-side refinement block that adaptively fuses encoder and decoder features.
Let $f_{\text{enc}}$ and $f_{\text{dec}}$ denote features from the encoder and decoder at a given UNet resolution. SCPM refines the decoder representation via

$$ f_{\text{refined}} = \gamma(f_{\text{enc}}) \odot \mathrm{Norm}(f_{\text{dec}}) + \beta(f_{\text{enc}}),$$

where $\gamma(\cdot)$ and $\beta(\cdot)$ are learned per-channel affine transformations (2-layer MLPs), and $\odot$ denotes element-wise multiplication. 

\begin{figure}[h]
    \centering
    \vspace{-3mm}
    \includegraphics[width=\textwidth]{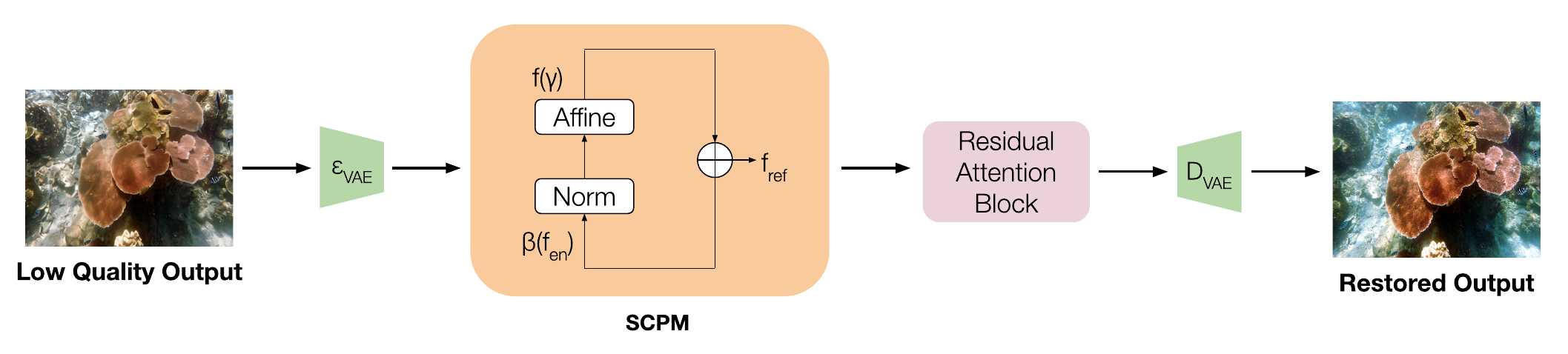}
    \caption{\textit{Semantic Content Preservation Module (SCPM).} Encoder features $f_{\text{enc}}$ are used to generate adaptive affine parameters $\gamma(f_{\text{enc}})$ and $\beta(f_{\text{enc}})$, which modulate normalized decoder features $\text{Norm}(f_{\text{dec}})$. The refined features $f_{\text{refined}}$ are then processed by residual and attention blocks before final decoding by $D_{\text{VAE}}$. This adaptive fusion preserves fine structures such as edges, textures, and small objects that are critical for scientific imaging tasks.}
    \label{fig:scpm_structure}
\end{figure}

The refined feature $f_{\text{refined}}$ is then passed through: (1) a residual convolutional block, (2) a lightweight self-attention layer, and (3) upsampled to the next scale. This design improves the preservation of edges, small textures, and domain-specific patterns, which are often blurred by diffusion bottlenecks. Appendix Figure \ref{fig:scpm_structure} shows the architecture of the SCPM.

\begin{figure}[h]
    \centering
    \vspace{-3mm}
    \includegraphics[width=0.9\textwidth]{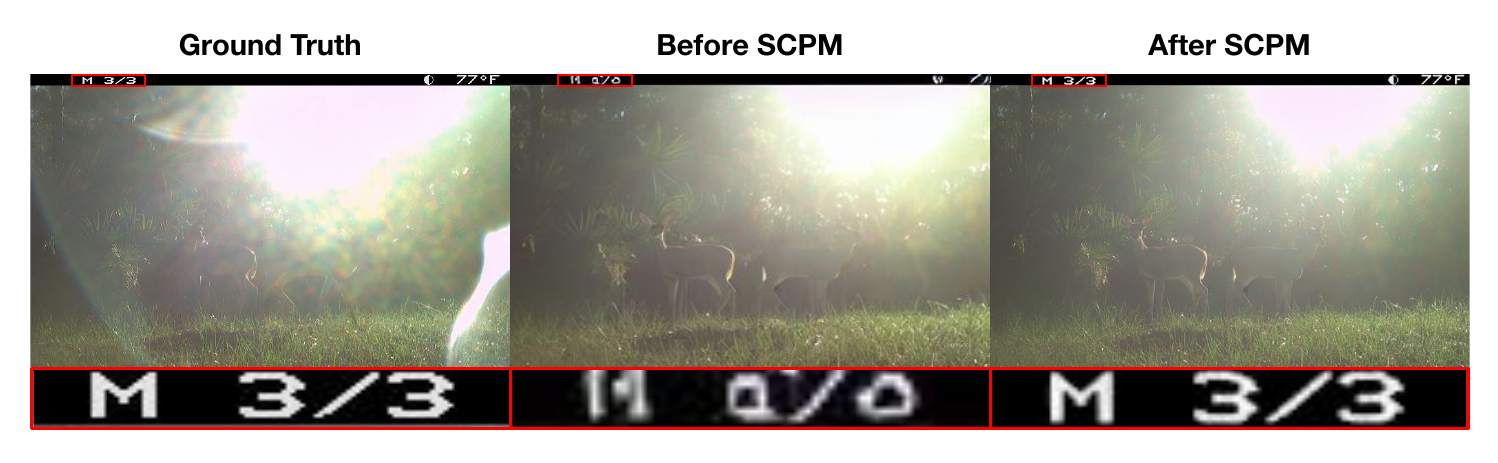}
    \caption{\textit{Effect of SCPM on restoration fidelity.} Without SCPM (middle), restoration reduces degradations but alters scene details, leading to blurred text/textures and distorted object boundaries. With SCPM (right), fine structures are preserved, maintaining fidelity to the ground truth (left). By reintroducing encoder features at the decoding stage, SCPM retains spatial cues that are often lost in the bottleneck representation. This cross-scale fusion constrains the decoder to stay faithful to the input structure, reducing hallucinations and over-smoothing while preserving fine details critical for scientific fidelity.}
    \label{fig:scpm_ablation}
\end{figure}

Quantitatively, removing SCPM drastically reduces MDB performance (see Table \ref{tab:ablations_supp}) and qualitatively, SCPM prevents content drift while recovering edges, textures, and small objects essential for downstream analysis (see Fig. \ref{fig:scpm_ablation}). 

\begin{table}[h]
\centering
\caption{\textit{Ablation on PRISM's SCPM.} This refinement module improves compound restoration fidelity on MDB.}
\label{tab:ablations_supp}
\scriptsize
\begin{tabular}{l|c c c}
\toprule
\textbf{} & \textbf{PSNR $\uparrow$} & \textbf{SSIM $\uparrow$} & \textbf{LPIPS $\downarrow$} \\
\midrule
After SCPM & \textbf{22.08} & \textbf{0.842} & \textbf{0.218} \\
Before SCPM & 16.65 & 0.654 & 0.566 \\
\bottomrule
\end{tabular}
\end{table}

\begin{figure}[h]
    \centering
    \vspace{-3mm}
    \includegraphics[width=\textwidth]{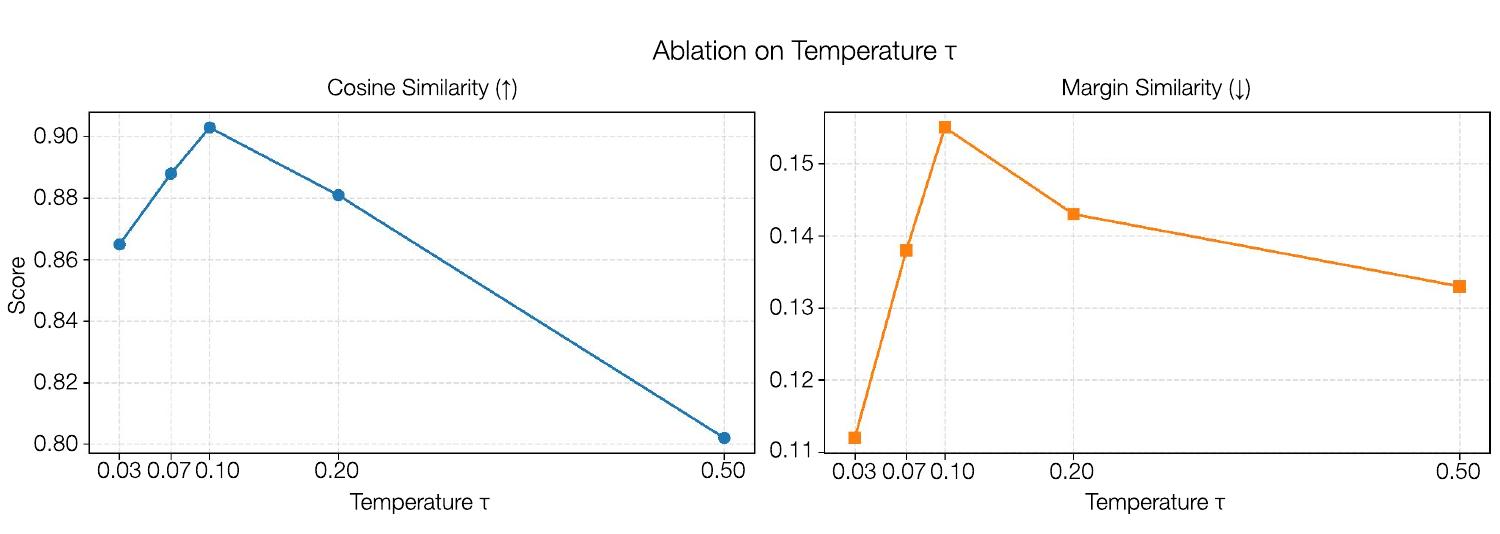}
    \caption{\textit{Ablation on temperature $\tau$.} for the contrastive objective. We show the mean cosine similarity between degraded and clean views (left) and the mean gap between positive and hardest-negative cosine (right). Results are means $\pm$ standard error margin over 3 seeds.}
    \label{fig:tau_ablation}
\end{figure}

\paragraph{Effect of Temperature $\tau$.}
We sweep $\tau \in \{0.03, 0.07, 0.10, 0.20, 0.50\}$ while keeping all other hyperparameters fixed. For each $\tau$, we train the embedding module and freeze it before training the diffusion backbone. Figure \ref{fig:tau_ablation} reports the average cosine similarity between degraded and clean views and the average difference between this positive similarity and the most confusable negative example (a large indicates better separation).

We observe a sweet spot at $\tau \approx 0.1$, which maximizes separation. Very low temperatures ($\tau=0.03$) over-emphasize hard negatives and reduce generalization; high temperatures ($\tau=0.5$) soften negatives excessively, collapsing cluster structure and harming retrieval/accuracy. We therefore set $\tau=0.1$ for all main results.

\paragraph{Role of Jaccard Re-weighting.}
Our setting involves \emph{multi-label} supervision, where each example is associated with a set of labels rather than a single class. For the contrastive objective, we require a similarity measure that reflects the \emph{degree of semantic overlap} between two such label sets. The Jaccard index is the canonical similarity measure for finite sets and is widely used in multi-label learning, retrieval, and segmentation \citep{lin2023effective, menke2025publication}). Unlike cosine or dot-product similarity between binary label vectors, the Jaccard index ignores shared negatives, which is critical in sparse label spaces, normalizes by the size of the union, ensuring comparability across examples with different label cardinalities, and provides a bounded and interpretable similarity in $[0,1]$, yielding smooth interpolation between positives and negatives. These properties make Jaccard an ideal choice for weighting sample pairs in our supervised contrastive loss.

We compare our Jaccard-based similarity weighting against the following alternatives:

\begin{itemize}
    \item \textbf{No re-weighting}: standard InfoNCE contrastive loss without contrastive reweighting,
    \item \textbf{Unweighted SupCon}: exact label matches treated as positives, all others as negatives,
    \item \textbf{Cosine similarity}: cosine between multi-hot label vectors,
    \item \textbf{Overlap coefficient}: $|y_i \cap y_j| / \min(|y_i|, |y_j|)$.
\end{itemize}

Each similarity score $s_{ij}$ is used in the same contrastive objective, with all training settings held fixed.

\begin{table}[h]
\centering
\scriptsize
\caption{\textit{Effect of similarity function on downstream performance on the main dataset.} Values show absolute performance change relative to the Jaccard-based loss (ours).}
\vspace{0.5em}
\begin{tabular}{lccc}
\toprule
\textbf{Similarity function} & {PSNR $\uparrow$} & {SSIM $\uparrow$} & LPIPS $\downarrow$\\
\midrule
None          & 19.52 & \textbf{0.734} & 0.154\\
Unweighted SupCon          & 20.63 & \textbf{0.799} & 0.388\\
Cosine (labels)            & 21.49 & 0.772 & 0.429 \\
Overlap coefficient        & 21.35 & 0.784 & 0.332 \\
\textbf{Jaccard (ours)}    & \textbf{22.08} & \textbf{0.842} & \textbf{0.218}\\
\bottomrule
\end{tabular}
\label{tab:jaccard_ablation_main}
\end{table}

Jaccard similarity yields the strongest performance. We suspect that alternatives such as cosine or the overlap coefficient systematically inflate similarity for high-cardinality label sets, reducing discriminativity. Unweighted SupCon performs worst because it discards all partial-overlap relationships, which are common in this dataset. Cosine similarity also correlates strongly with label count (Spearman $\rho = 0.42$), whereas Jaccard minimizes this dependency ($\rho = 0.07$), ensuring that similarity reflects the \emph{fraction of shared labels} instead of the absolute number of labels present. Table \ref{tab:jaccard_ablation_main} also shows that without any re-weighting in the contrastive loss (using a standard InfoNCE-based loss), overall performance decreases drastically.

Many sample pairs share only 1-2 labels. Jaccard naturally assigns intermediate similarity values in these cases, enabling ``soft positives'' that improve representation learning. We expect that the overlap coefficient overestimates similarity for such pairs, while Unweighted SupCon ignores them entirely.

\paragraph{Role of Distortion Complexity}

Here we provide empirical evidence and theoretical motivation behind training on up to three distortions. To directly evaluate this design choice, we sweep the maximum number of distortions used during training over $N \in \{1,2,3,4\}$ while holding all other factors constant (same set of clean images, same set of augmentations, same CLIP fine-tuning, same diffusion backbone). For each setting, we re-train PRISM from scratch and evaluate restoration fidelity (PSNR, SSSM, and LPIPS) on \textit{images from MDB that are degraded by a single distortion}. We also evaluate training stability by the variance of the validation loss across epochs. Lastly, we retrain and evaluate the accuracy of a lightweight multi-label classifier on identifying the distortions present from the CLIP image encoder. Table~\ref{tab:n_sweep_full} summarizes results across all distortion-count settings.

\begin{table}[h]
\centering
\caption{{\textit{Impact of maximum distortion count ($N$) used in the training set.}
Training on more than three simultaneous degradations reduces restoration fidelity, harms distortion classification accuracy, and destabilizes prompt-following behavior.}}
\label{tab:n_sweep_full}
\scriptsize
\begin{tabular}{c|c c c|c|c}
\toprule
\multirow{2}{*}{$\mathbf{N}$} &
\multicolumn{3}{c|}{\textbf{Restoration Fidelity}} &
\textbf{Training} &
\textbf{Classifier}\\
& PSNR $\uparrow$ & SSIM $\uparrow$ & LPIPS $\downarrow$ &
Stability $\downarrow$ &
F1 Score $\uparrow$\\
\midrule
1 & 24.35 & 0.942 & 0.112 & 0.014 & 0.91 \\
2 & 20.65 & 0.923 & 0.126 & 0.018 & 0.88 \\
\hline
3 & 18.73 & 0.842 & 0.218 & 0.022 & 0.87 \\
\hline
4 & 16.98 & 0.741 & 0.401 & 0.047 & 0.61\\
\bottomrule
\end{tabular}
\end{table}

From this sweep, we observe diminishing returns and eventual collapse beyond $N=3$. Increasing $N$ from 1 to 3 expands the compositional coverage of the data augmentation pipeline, enabling the model to learn meaningful interactions among common distortions (e.g., blur+noise, haze+color shift). When 4 or 5 distortions are applied simultaneously, synthetic degradations become so severe that the underlying semantic content is ambiguously recoverable, hurting both restoration fidelity and distortion-classification accuracy (F1 drops from 0.87 to 0.61).  This aligns with the intuition that supervision becomes unreliable as compound degradations multiply. Thus, $N=3$ maximizes the trade-off between representing realistic compound degradations (seen in underwater, microscopy, remote sensing, and camera-trap imagery) and maintaining well-posed, interpretable supervision for restoration.

\paragraph{Role of Partial and Negative Prompts.}  
Training with partial prompts (requesting removal of only a subset of degradations applied to an image) and negative prompts (explicitly requesting removal of degradations not present) enforces controllability. Without these cases, the model tends to over-restore, indiscriminately removing everything it detects. To evaluate this, we compute prompt faithfulness: for each prompt, we compare the predicted degradation labels before and after restoration against the degradations specified in the prompt. A restoration is counted as faithful if all requested degradations are removed while non-requested degradations are preserved. As shown in Table \ref{tab:ablations_prompts}, including partial and negative prompts during training improves prompt faithfulness.

\paragraph{Role of Prompt Diversity.}  
We generate multiple natural-language variants for each distortion type (e.g., ``remove haze,'' ``clear atmospheric fog''). Limiting training to a fixed prompt format (``remove the effects of haze'') only improves prompt faithfulness by 0.4\%. Considering the tradeoff between accuracy and usability, we conclude that the benefits of linguistic variability outweigh this minor change in performance.

\begin{table}[h]
\centering
\caption{\textit{Effect of partial and negative prompts.} Including these improves prompt faithfulness (measured as proportion of outputs correctly following instructions).}
\label{tab:ablations_prompts}
\scriptsize
\begin{tabular}{l|c}
\toprule
\textbf{\hspace{3mm}Training Setting\hspace{3mm}} & {\hspace{3mm}Prompt Faithfulness $\uparrow$\hspace{3mm}} \\
\midrule
w/o Partial or Negative Prompts & 61.4\% \\
With Partial Prompts Only & 78.9\% \\
With Negative Prompts Only & 85.9\% \\
w/o Prompt Variation & 88.1\% \\
With Partial + Negative Prompts and Prompt Variation (ours) & 87.7\% \\
\bottomrule
\end{tabular}
\end{table}

\paragraph{Sensitivity to Prompting Style.}
\label{sec:automation}

To clarify that PRISM does not require human labor at scale, we compare three prompting modes: (1) fixed manual prompts of the form ``remove the effects of x, y, and z,'' (2) automated prompts produced by PRISM's multi-label distortion classifier, and (3) free-form user prompts paraphrased via GPT-4 (sampled from the pre-existing pool of variants of the fixed-prompt) to simulate real-world usage.

\begin{table}[h]
\centering
\caption{Comparison of prompting modes on MDB. 
Automated restoration achieves 85--95\% of fixed prompting performance.}
\vspace{0.2cm}
\scriptsize
\begin{tabular}{l c c c}
\toprule
\textbf{Method} & \textbf{PSNR $\uparrow$} & \textbf{SSIM $\uparrow$} & \textbf{LPIPS $\downarrow$} \\
\midrule
Fixed Prompt       & 22.08 & {0.842} & 0.218 \\
Automated Prompt   & 20.84 & 0.824          & 0.229 \\
Free-form Prompt   & 20.81 & 0.819          & 0.231 \\
\bottomrule
\end{tabular}
\label{tab:automation_modes}
\end{table}

Automation provides near-manual performance while eliminating human specification entirely.
This confirms that PRISM is not dependent on expert labor for large-scale deployment.
Free-form prompting performs similarly, indicating that PRISM tolerates paraphrasing and user variation.

We evaluate PRISM’s distortion classifier on the MDB test set spanning 14 primitive distortions and their synthetic combinations. The classifier achieves 0.94 accuracy on this benchmark, reliably identifying both individual degradations and multi-distortion mixtures. This high classification fidelity helps explain the strong automated restoration results reported in Table~\ref{tab:automation_modes}. When uncertain, the classifier tends to predict supersets of the true distortions (e.g., “blur + haze’’ instead of “haze’’), a desirable bias in restoration where omission is riskier than mild over-specification.

However, MDB is a synthetic benchmark with cleanly defined distortion categories. It does not capture the open-ended, overlapping, and often ambiguous distortion mixtures found in real imagery. As a result, MDB accuracy cannot directly evaluate real-world generalization, and we expect classifier accuracy to be lower on real compositional degradations, where distortions are less separable and category boundaries are not well-defined.

\paragraph{Sensitivity to Prompt Granularity and Accuracy}
\label{sec:prompt_granularity}

To explicitly quantify how different forms of prompt specification influence PRISM’s behavior, 
we perform a controlled prompt-perturbation study on the MDB.
For all experiments in this subsection, we use the same MDB test set validation images, 
ensuring that each prompt type is evaluated on \emph{identical} inputs and degradation patterns.

For every image $I$ with ground-truth distortion set $\mathcal{D}$, we construct four prompt variants using GPT-4:

\begin{enumerate}

    \item \textbf{Composite semantic:} A single combined instruction describing the same distortion set 
    (e.g., ``remove the effects of blur and haze''). These prompts are generated using a fixed template to avoid stylistic confounds. These prompts are generated using the form ``remove the effects of x, y, and z.'' 

    \item \textbf{Over-specific:} Prompts augmented with redundant adjectives or intensity qualifiers 
    (e.g., ``remove mild Gaussian blur and strong chromatic haze''). For each set of distortions, we query GPT-4 with ``Write 10 restoration instructions that tells a model to remove the following distortions x, y, and z. For each distortion, add one or two descriptive modifiers (e.g., mild, severe, colored, high-frequency, patchy).''

    \item \textbf{Under-specified:} Prompts that target broad categories of restoration
    (e.g., true set: snow, prompt: ``remove occlusions''). For each set of distortions, we query GPT-4 with ``Write 10 broad, high-level restoration instructions that could apply to the following distortions x, y, and z. Use a generic category name such as 'occlusions' or 'artifacts.' Do not mention specific distortion types.''

    \item \textbf{Incorrect / negative:} Prompts specifying distortions \emph{not} present in $\mathcal{D}$, sampled uniformly from the remaining distortion types 
    (e.g., true set: blur+haze, prompt: ``remove rain''). We used the fixed format ``remove the effects of x, y, and z.''
\end{enumerate}

For each prompt variant, we encode the instruction using the {frozen CLIP text encoder} 
from our distortion-aware contrastive training stage, and condition the latent diffusion model 
with this embedding. Image restoration is run using the same sampling schedule and model weights across all conditions.

We report PSNR, LPIPS, prompt faithfulness, and the cosine similarity between the CLIP 
text embedding of the prompt and the CLIP image embedding of the degraded input
(after applying the distortion-aware fine-tuning). 
The latter serves as a measure of \emph{latent alignment} between the user instruction 
and the distortion evidence present in the image.

\begin{table}[h]
\centering
\caption{Ablation on prompt granularity and accuracy. 
PRISM is robust to linguistic variations but sensitive to the correctness of the distortion set.}
\vspace{0.2cm}
\scriptsize
\begin{tabular}{l c c c c}
\toprule
Prompt Type & PSNR $\uparrow$ & LPIPS $\downarrow$ & Faithfulness $\uparrow$ & Alignment $\uparrow$ \\
\midrule
Composite semantic & {22.08} & {0.218} & {0.98} & {0.93} \\
Over-specific & 22.01 & 0.225 & 0.96 & 0.89 \\
Under-specified & 20.30 & 0.241 & 0.78 & 0.70 \\
Incorrect / negative & 21.39 & 0.247 & 0.55 & 0.53 \\
\bottomrule
\end{tabular}
\label{tab:prompt_granularity}
\end{table}

Results show that PRISM is invariant to changes in linguistic style (primitive-only vs.\ composite vs.\ over-specific), 
but is sensitive to whether the prompt specifies the correct \emph{set of underlying distortions}. 
Importantly, even when prompts are incorrect or incomplete, the image-driven CLIP embedding 
pulls the joint embedding back toward an appropriate region of the latent space, limiting degradation.
This supports the claim that PRISM is robust and image-conditioned, rather than overly dependent on prompt wording.

\begin{figure}[h]
\centering
\includegraphics[width=\textwidth]{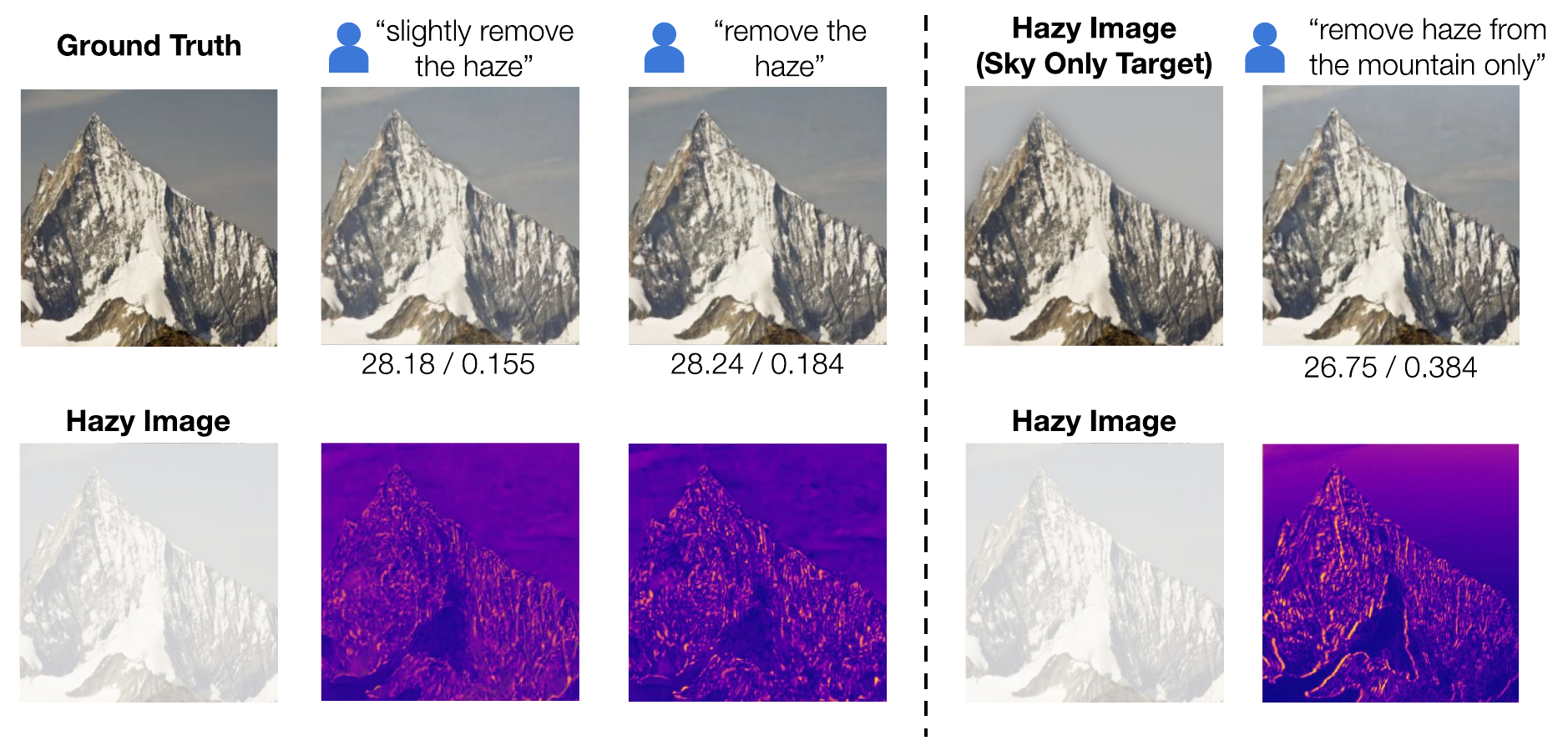}
\caption{{\textit{Responses to intensity- and location-aware prompts.} PRISM shows qualitative variability with different prompt phrasings: milder restoration tends to occur under “slightly remove the haze,” while stronger dehazing appears under “remove the haze.” Spatial specificity remains limited, and the model generally applies restoration globally even when local changes are requested (e.g., “remove haze from sky”), likely reflecting its training on global augmentations. Bottom row: heatmaps visualize pixel-wise error, and PSNR/LPIPS values are reported below.}}
\label{fig:local_control}
\end{figure}

\vspace{2mm}
\paragraph{Sensitivity to Local and Intensity-Aware Prompts.}
We examine PRISM’s behavior when given restoration instructions that reference either the strength (e.g., “slightly remove the haze”) or the location (e.g., “remove haze from sky”) of degradation removal. Although PRISM was not explicitly trained on scalar intensity levels or spatial masks, we observe that it can exhibit qualitative differences in response to intensity-related modifiers, with prompts such as “slightly” sometimes resulting in milder restorations (see Fig.~\ref{fig:local_control}). In contrast, spatially localized prompts are interpreted less consistently. This is unsurprising given that our data augmentation pipeline introduces degradations uniformly across the entire image, meaning the model is primarily exposed to globally applied distortions during training. Future work incorporating localized augmentations and corresponding context-specific prompts may help improve performance in this setting.

\paragraph{Cost and Latency Analysis}

In this section, we benchmark efficiency, memory usage, and latency under standardized conditions to evaluate the practical deployability of PRISM relative to both lightweight restoration models and state-of-the-art diffusion-based baselines.

All evaluations were conducted on a single NVIDIA A100 (40GB) GPU using a fixed input resolution of $256 \times 256$. We report FLOPs (via the `thop` library), peak GPU memory usage (via PyTorch profiling tools), and average per-image latency (ms). All models were evaluated using identical batch size, mixed-precision (fp16), and PyTorch’s benchmarking mode for consistent measurement.

\begin{table}[h]
\centering
\caption{\textit{Efficiency comparison under standardized conditions.} Despite added controllability and SCPM modules, PRISM matches the runtime and memory footprint of strong diffusion-based baselines.}
\label{tab:latency_cost}
\scriptsize
\begin{tabular}{l|c c c}
\toprule
\textbf{Method} & \textbf{FLOPs (G)} & \textbf{Memory (GB)} & \textbf{Latency (s $\downarrow$)} \\
\midrule
AirNet~\citep{airnet} & 46 & 2.1 & 0.3 \\
Restormer$_A$~\citep{restormer} & 118 & 4.6 & 0.5 \\
NAFNet$_A$~\citep{nafnet} & 104 & 4.2 & 0.5 \\
OneRestore~\citep{guo2024onerestore} & 136 & 5.8 & 0.6 \\
PromptIR~\citep{potlapalli2023promptir} & 128 & 5.4 & 0.6 \\
\midrule
DiffPlugin~\citep{liu2024diff} & 145 & 6.5 & 2.63 \\
MPerceiver~\citep{ai2024multimodal} & 132 & 5.9 & 1.98 \\
AutoDIR~\citep{jiang2024autodir} & 138 & 6.0 & 2.24 \\
\textbf{PRISM (manual)} & 141 & 6.1 & 2.55 \\
\textbf{PRISM (automated)} & 141 & 6.1 & 2.87 \\
\bottomrule
\end{tabular}
\end{table}

As expected, lightweight encoder–decoder models such as AirNet offer the fastest throughput and lowest memory usage, but struggle with generalization and complex distortions as shown in the main text of the paper. Transformer-based or prompt-conditioned models (Restormer, NAFNet, PromptIR) increase compute due to deeper backbones and multi-branch design.

Among diffusion-based models, PRISM operates at a similar computational cost to leading baselines (DiffPlugin, AutoDIR, MPerceiver), with a modest increase in FLOPs and memory due to our SCPM conditioning module. Importantly, this added capability yields improved downstream fidelity (Tables 2, 5, and 9 in main text and Appendix). The distortion classifier used in automated restoration incurs minimal additional cost compared to manual restoration. We believe this demonstrates that prompt-guided, compound-aware restoration is achievable without compromising deployability—making PRISM well-suited for integration into scientific workflows that require interpretability and modularity alongside high-quality results.

\section{Additional Figures}
\label{sec:add_figures}

\begin{figure}[h]
    \centering
    \includegraphics[width=0.75\textwidth]{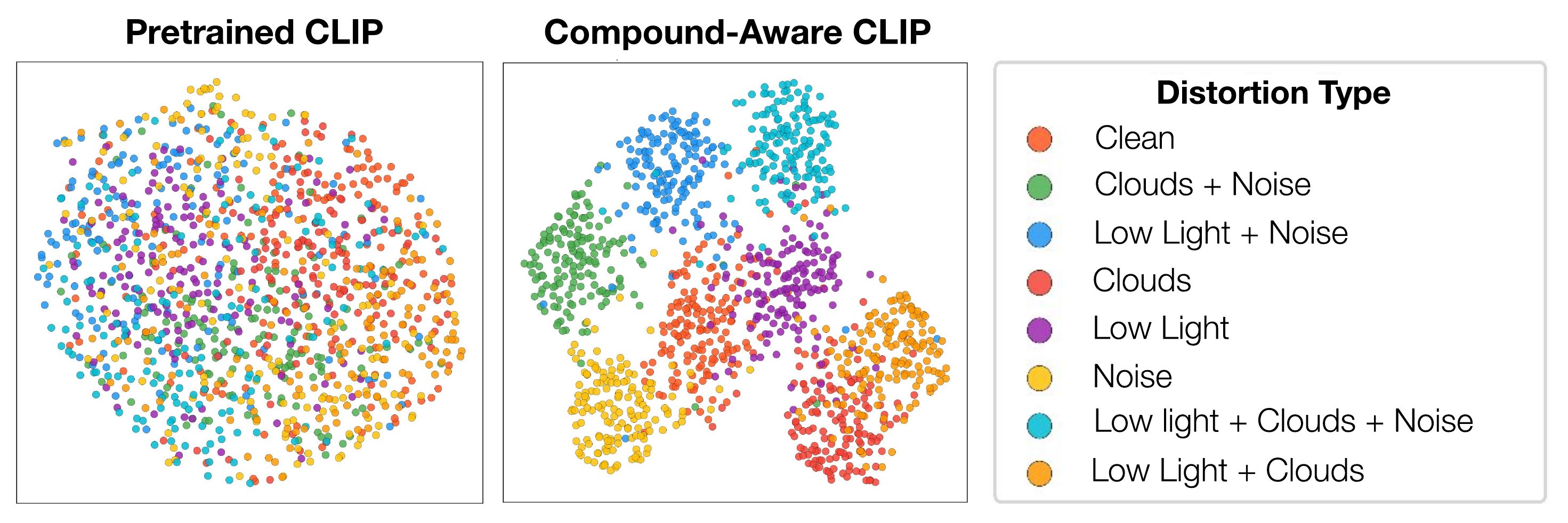}
    \caption{\textit{Contrastive disentanglement of distortion embeddings.} UMAP projections of $f(I_{\text{dist}})$ from 10K samples in the MDB, across a subset of distortion classes. Left: CLIP entangles distortions with semantics. Right: Our weighted contrastive loss achieves clear separation while aligning compounds with their primitives. Overall, without contrastive disentanglement, embeddings of compound degradations collapse into unrelated regions, forcing the model to treat them as unseen categories. This can lead to artifacts or overcorrection.}
    \label{fig:latent}
\end{figure}

\begin{figure}[h]
    \centering
    \includegraphics[width=0.7\textwidth]{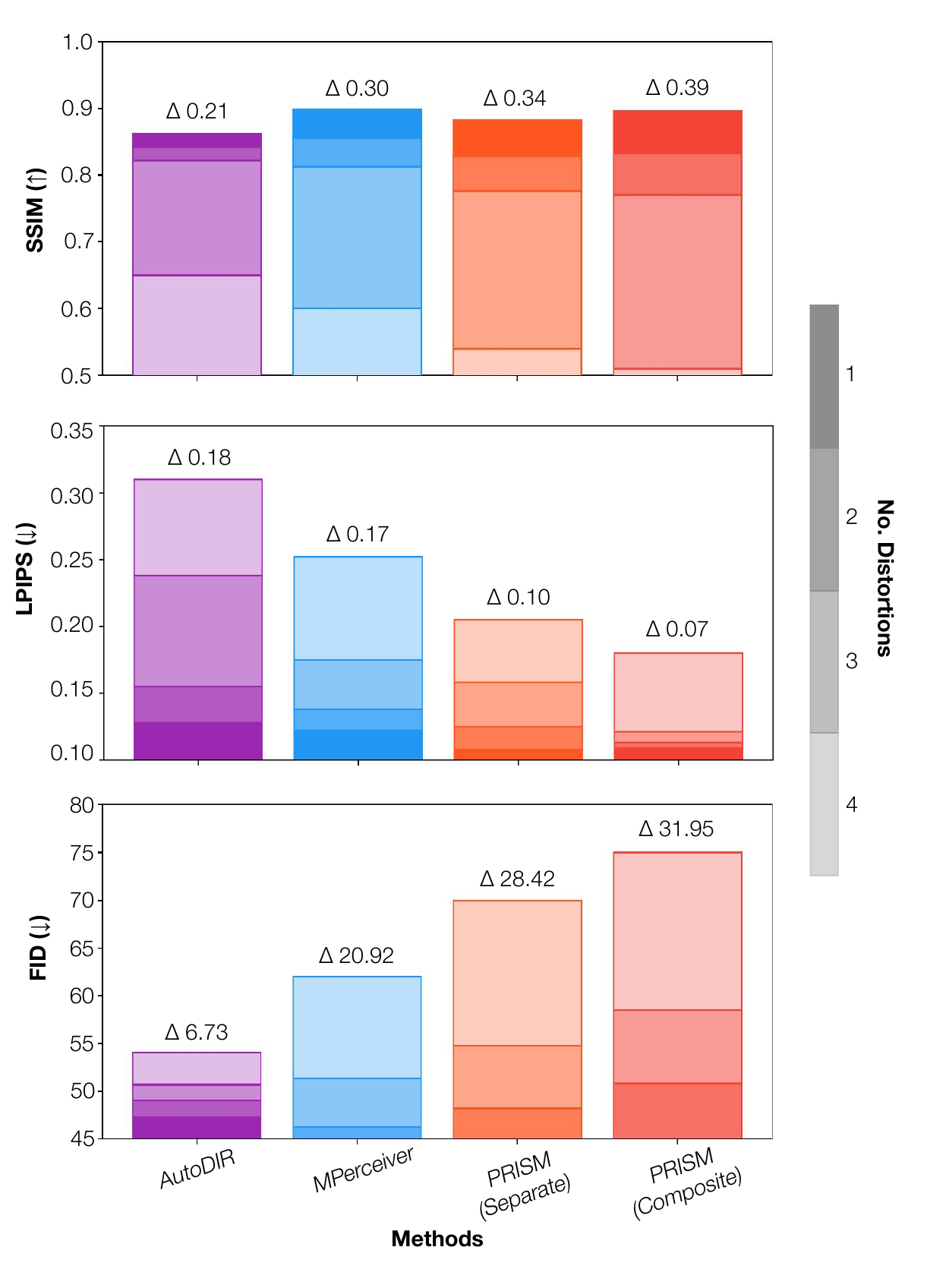}
    \caption{\textit{PRISM trained on composite examples scales best with the number of distortions.} This outperforms PRISM trained on each degradation separately as well as comparable baselines (MPerceiver and AutoDIR), emphasized by the $\Delta$ (change in performance across test images with 1 vs. 4 distortions) above each bar.}
    \label{fig:composite_app}
\end{figure}

\begin{figure}[h]
    \centering
    \includegraphics[width=0.4\textwidth]{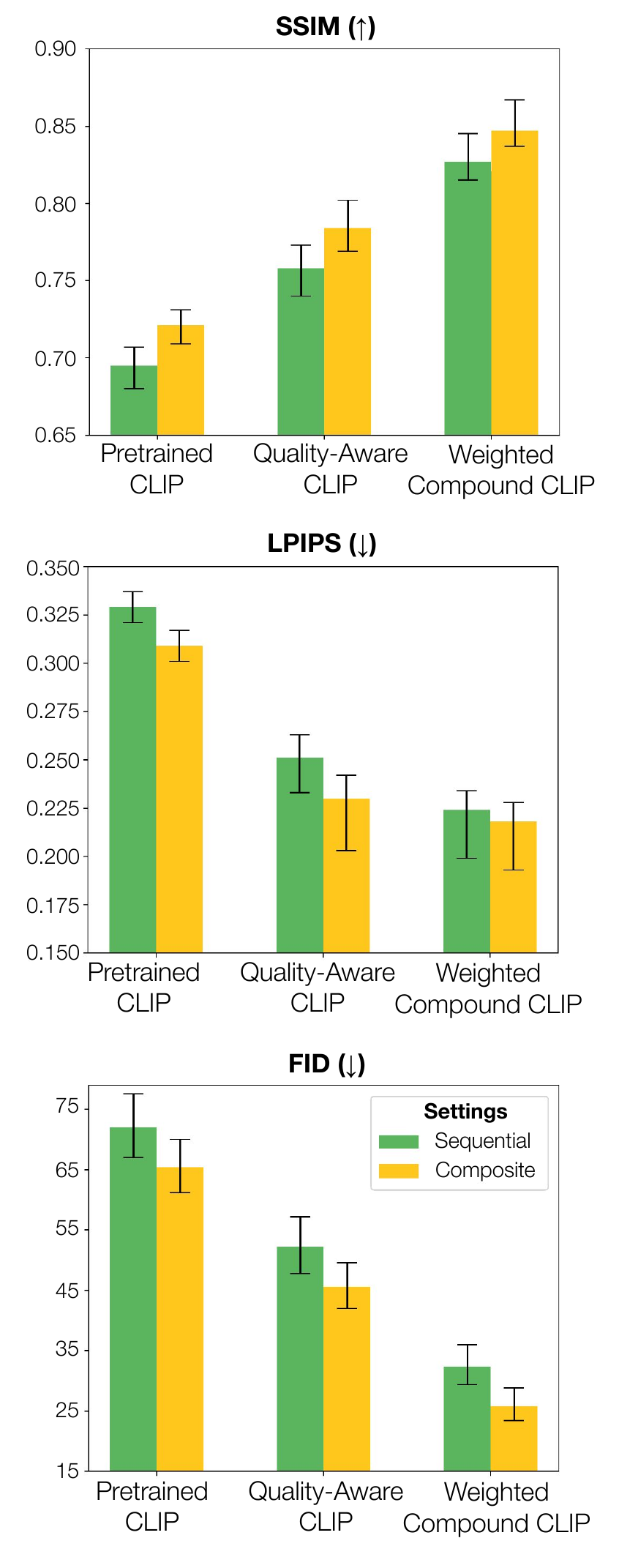}
    \caption{\textit{Latent disentanglement of distortion types enables faithful stepwise and single-shot restoration.} The contrastive loss closes the gap between sequential and composite prompting.}
    \label{fig:loss_app}
\end{figure}

\begin{figure}[h]  
    \vspace{-3mm}
    \centering 
    \includegraphics[width=\textwidth]{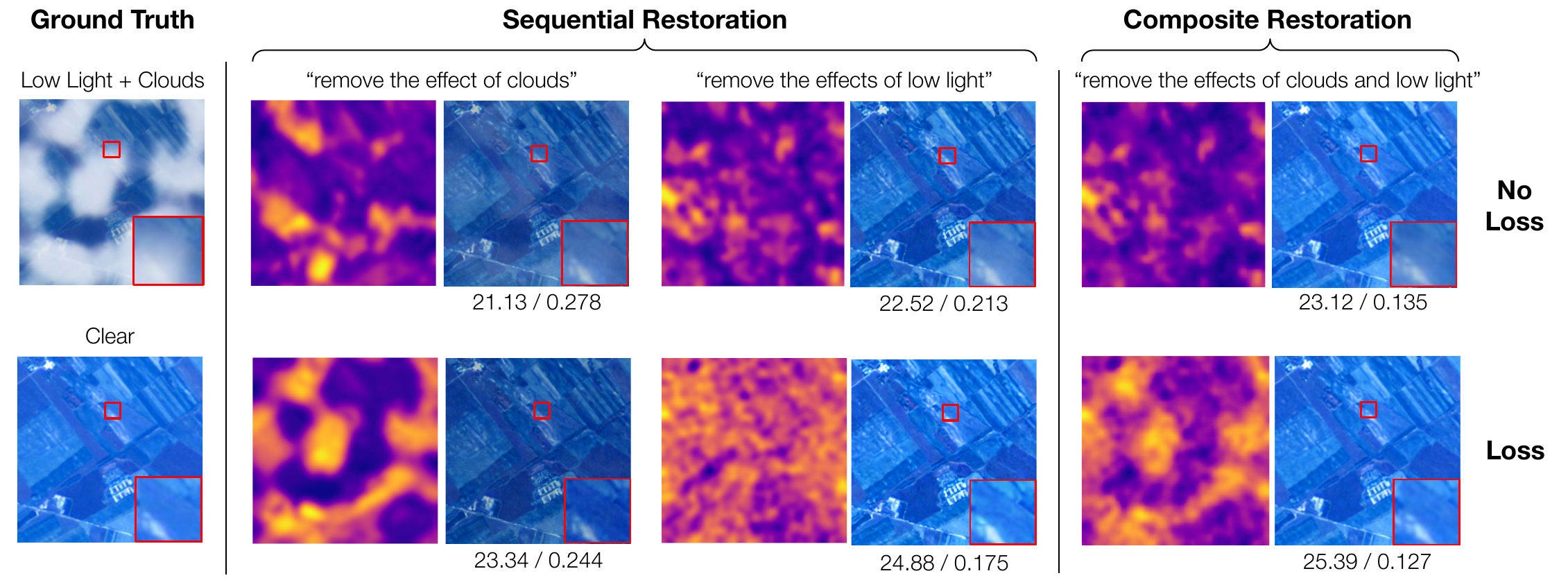} 
    \caption{\textit{Contrastive disentanglement of distortions helps separate distortions from each other and from semantic content, enabling higher-fidelity sequential and composite restoration.} Cross-attention maps (left of each output) show how the model attends to distortions. Without PRISM’s contrastive disentanglement (top), sequential restoration preserves artifacts and fails to isolate degradations. With the loss (bottom), embeddings cleanly separate distortions (e.g., clouds vs. low light). This separation not only prevents distortion types from interfering with one another, improving sequential restoration by reducing error accumulation, but also enables the model to accurately target and remove multiple degradations simultaneously, as seen in the composite restoration outputs. We report PSNR/LPIPS metric values below each output.}
    \label{fig:attention}
    \vspace{-3mm}
\end{figure}

\begin{figure}[h]
    \centering
    \includegraphics[width=\textwidth]{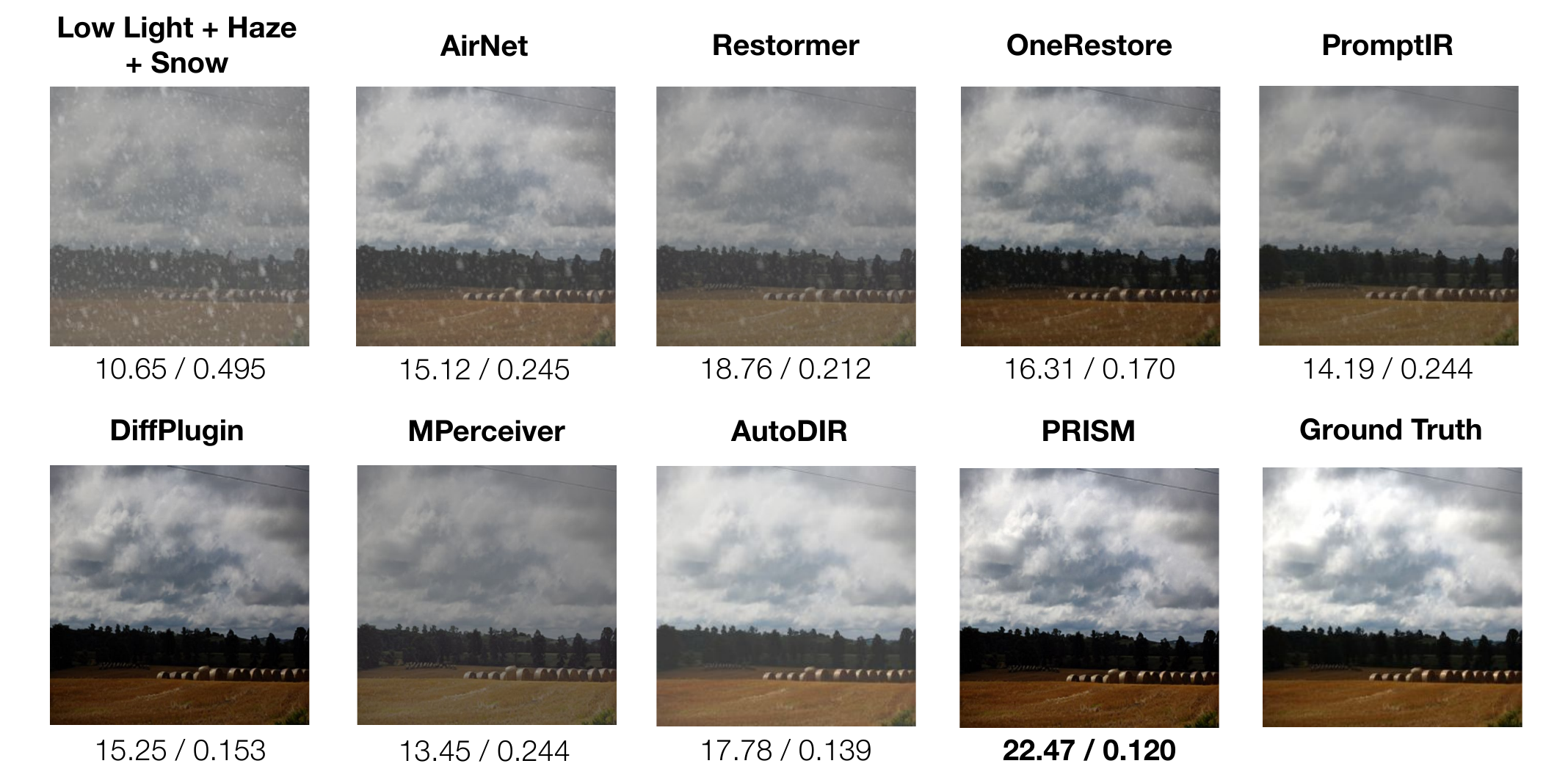}
    \caption{\textit{Qualitative outputs on the Mixed Degradations Benchmark (MDB).} Example of a low-light + haze + snow composite evaluated across baselines. We report (PSNR/LPIPS) below, with the best results in \textbf{bold}. While prior methods reduce some degradations, they leave residual haze (AirNet, PromptIR), oversmooth texture (Restormer, MPerceiver), or introduce artifacts from over-correction (OneRestore, AutoDIR, DiffPlugin). PRISM produces the most faithful reconstruction, recovering both sky and foreground with minimal artifacts, closely matching the ground truth. This illustrates the strength of compositional latent disentanglement: PRISM not only removes multiple degradations simultaneously but also resists the tendency to over-restore, yielding outputs that are both high fidelity and scientifically faithful.}
    \label{fig:mdb_qualitative}
\end{figure}

\begin{figure}[h]
    \vspace{-3mm}
    \centering
    \includegraphics[width=0.9\textwidth]{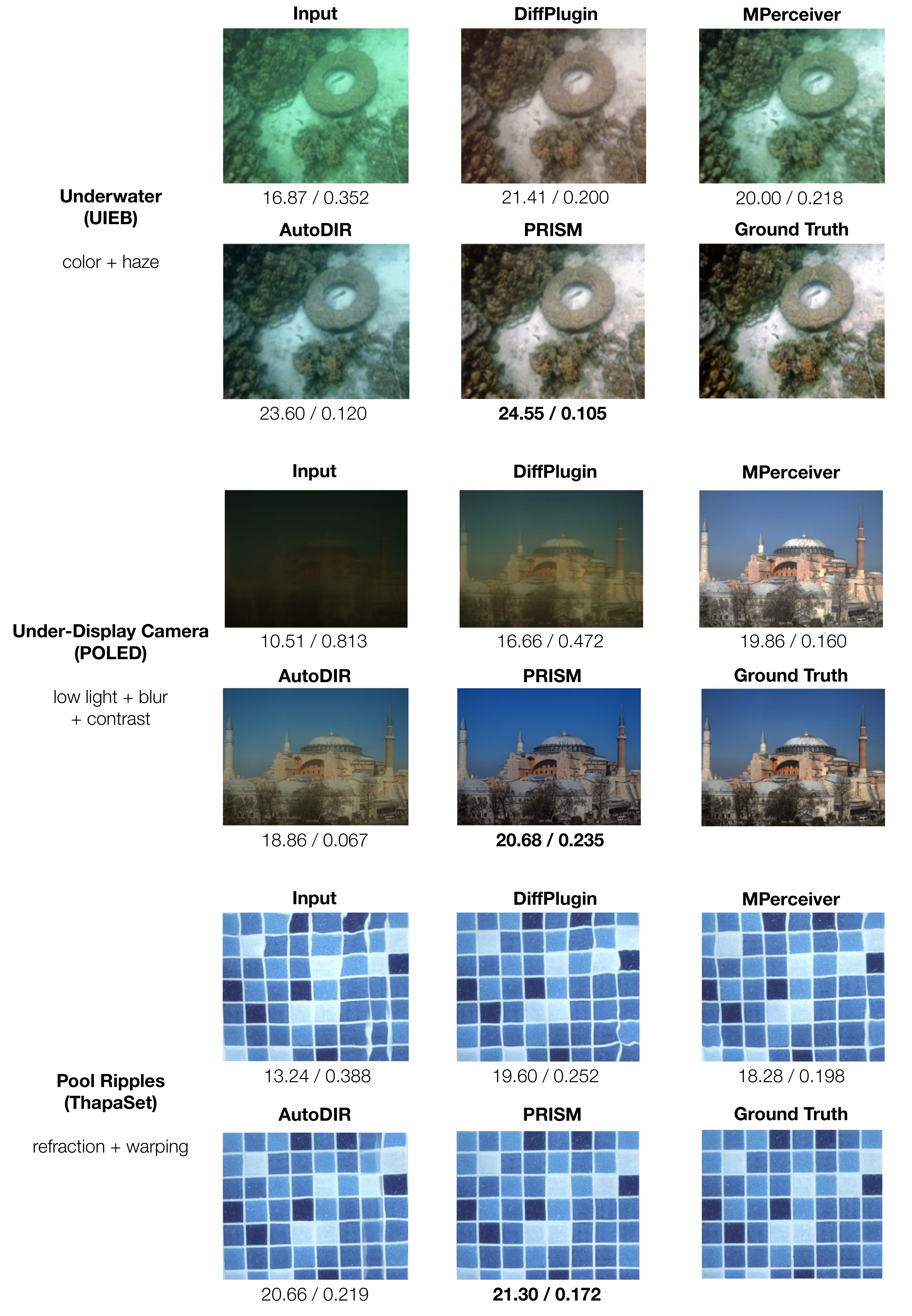}
    \caption{\textit{PRISM best removes unseen compound degradations.} PRISM restores real-world images with degradations outside its training set in underwater imagery, under-display camera images, and fluid lensing. In all cases, it produces faithful restorations that most closely match the ground truth, showing strong single-shot generalization compared to similar diffusion baselines. We report PSNR/LPIPS metric values below, with the best results in \textbf{bold}.}
    \label{fig:unseen}

    \vspace{-3mm}
\end{figure}

\label{sec:downstream_app}
\begin{figure}[h]  
    \vspace{-3mm}
    \centering 
    \includegraphics[width=\textwidth]{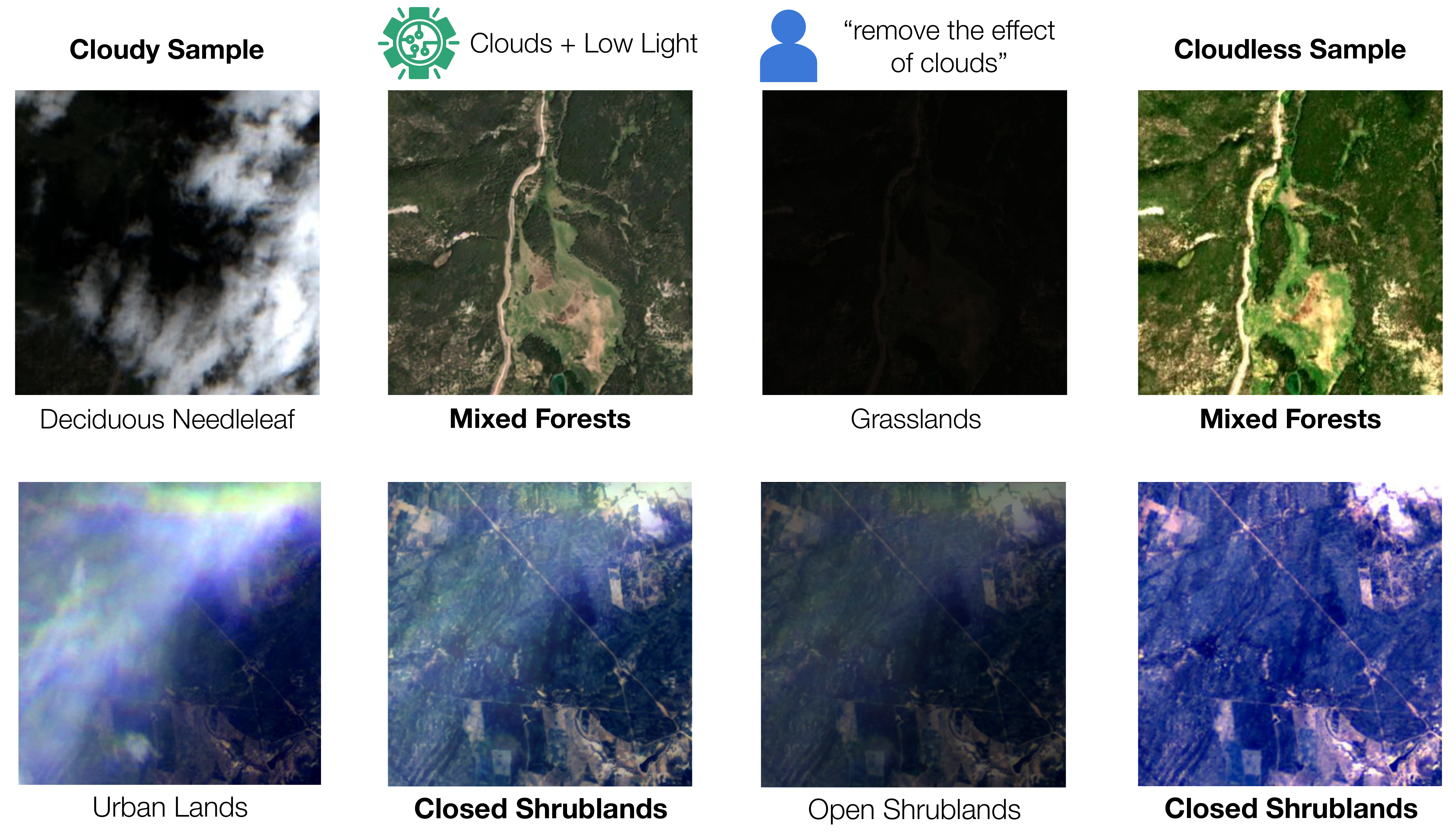} 
    \caption{\textit{Remote sensing classification under cloud occlusions requires full composite restoration.} In this Sentinel-2 example, removing only clouds (middle-left) reveals incomplete information and leads to a misclassification. Full composite restoration (middle-right), correcting both clouds and low light, recovers the underlying landscape with high fidelity and matches the ground-truth class, in \textbf{bold}. }
    \label{fig:clouds} 
    \vspace{-3mm}
\end{figure}

\begin{figure}[h]  
    \centering 
    \includegraphics[width=0.85\textwidth]{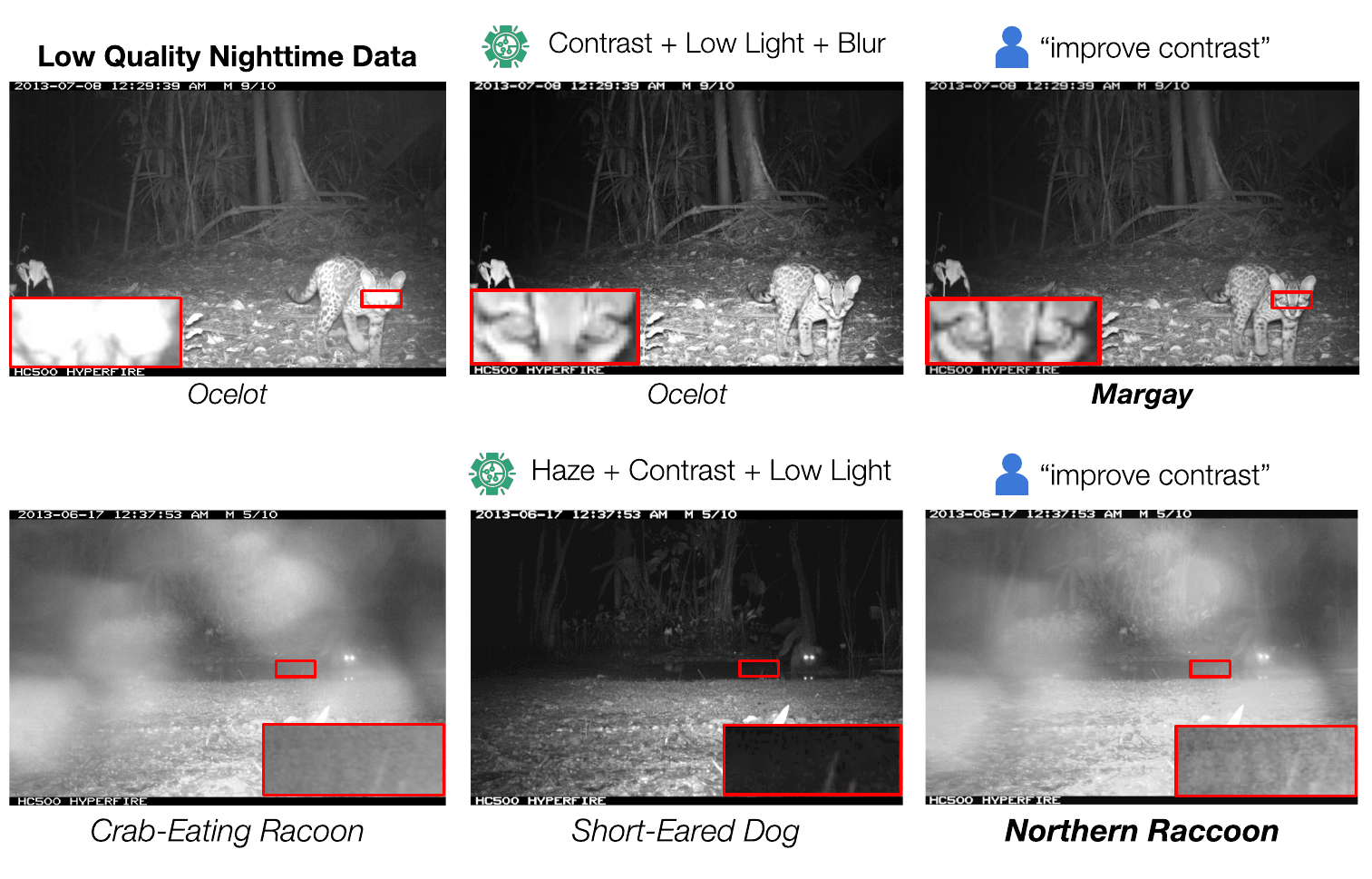} 
    \caption{\textit{Selective restoration helps with camera trap classification under compound nighttime degradations.} On the Rooftop Cityscapes dataset, frames suffer from haze and low-light conditions. Only improving contrast aids recognition of nocturnal species, while over-restoration (e.g., removing haze) can alter image content, obscure subtle texture cues, or introduce artifacts that mislead classification—sometimes even changing the perceived species. We \textbf{bold} the classification outputs that matches expert-provided labels.}
    \label{fig:species} 
\end{figure}

\begin{figure}[h] 
    \vspace{-3mm}
    \centering 
    \includegraphics[width=\textwidth]{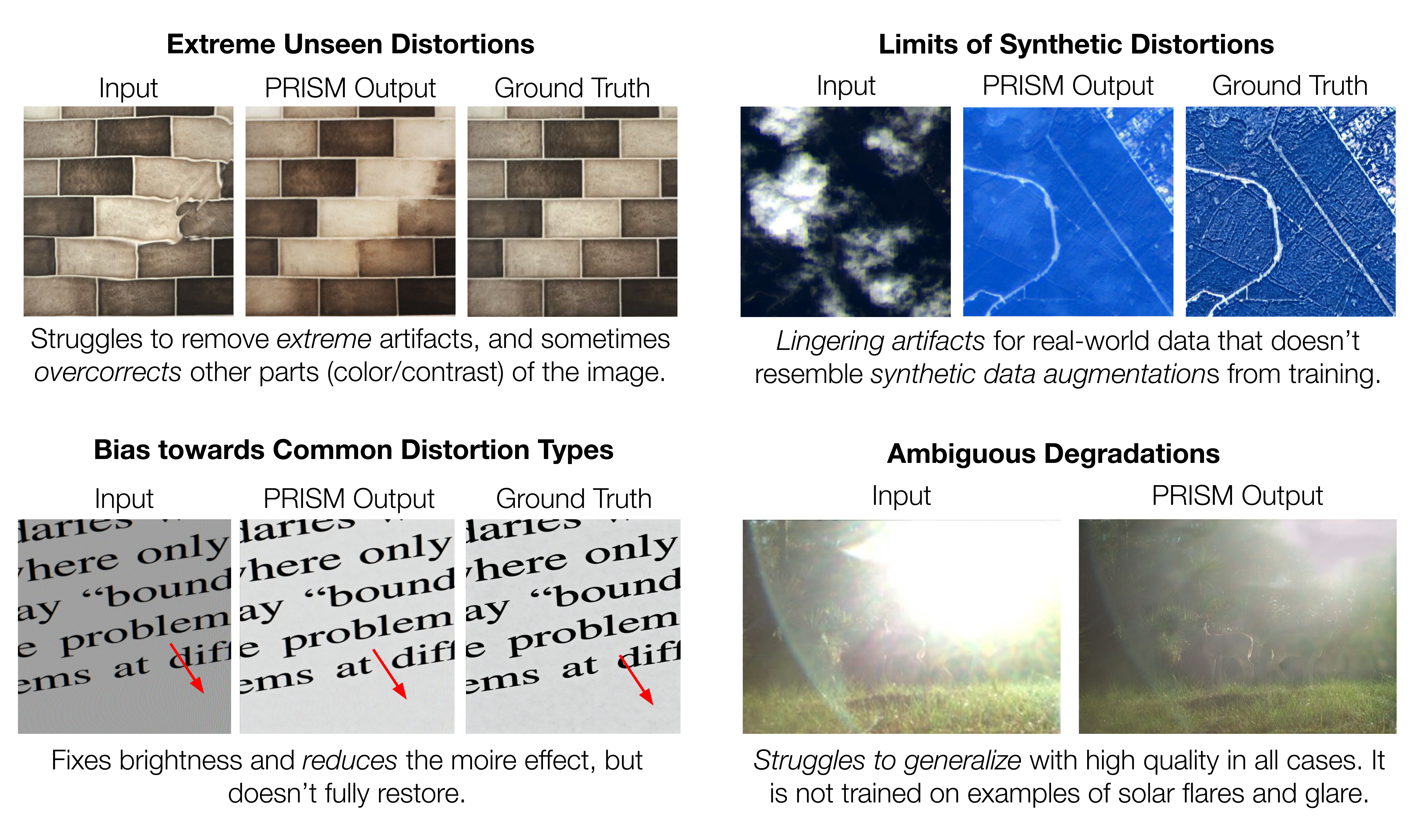} 
    \caption{\textit{Failure modes of PRISM on challenging degradations.} Top-left: Extreme unseen distortions cause incomplete restoration and overcorrection of color/contrast. Top-right: Overfitting to synthetic distortions leaves lingering artifacts when applied to real data that diverges from training augmentations. Bottom-left: Overfitting to common distortions partially reduces moire but fails to fully restore fine details. Bottom-right: Ambiguous degradations (e.g., solar flares, glare) remain difficult to generalize without explicit training examples.} 
    \label{fig:failure_modes} 
    \vspace{-3mm}
\end{figure}

\begin{figure}[h!]
    \centering
    \vspace{-3mm}
    \includegraphics[width=0.8\textwidth]{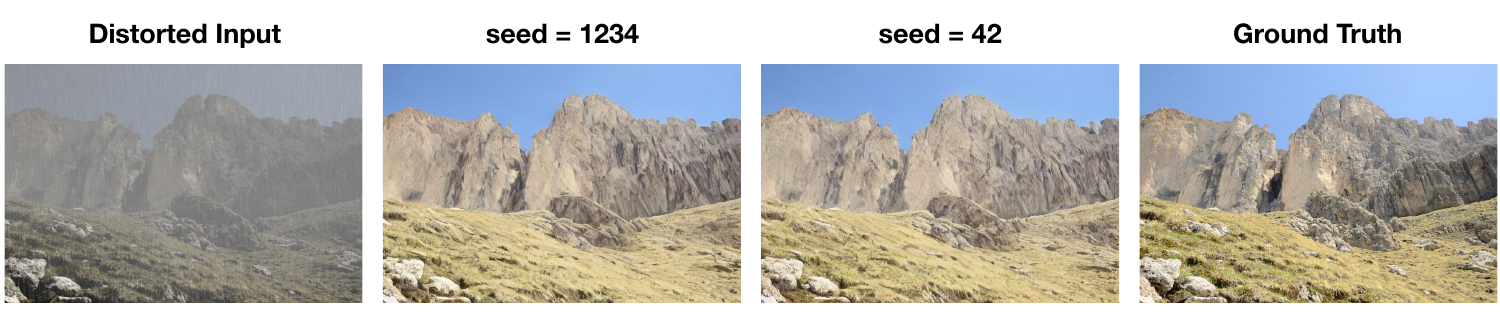}
    \caption{\textit{Qualitative impact of random seed and stochasticity on restoration outcomes.} Different seeds produce slightly varied outputs, reflecting both diffusion sampling variability and embedding initialization. While global structure remains stable, fine details may differ, underscoring the importance of evaluating consistency across multiple runs.}
    \label{fig:stochasticity}
\end{figure}

\end{document}